\documentclass[a4paper,10pt]{article}

\usepackage{amsmath, amssymb, graphicx, fancyhdr, geometry, titlesec, hyperref, booktabs}

\usepackage{scrextend}
\usepackage{float}
\deffootnote{3mm}{1.6em}{\thefootnotemark.\enskip}

\newcommand{\AffiliationText}{}
\newcommand{\EmailText}{}

\newcommand{\affiliation}[1]{\renewcommand{\AffiliationText}{#1}}
\newcommand{\email}[1]{\renewcommand{\EmailText}{#1}}

\makeatletter
\renewcommand{\maketitle}{
    \begin{center}
        \vspace*{1cm}
        {\LARGE \bfseries \@title \par}
        \vspace{1em}
        {\large \@author \par}
        \vspace{1em}
    \end{center}
    \ifx\AffiliationText\empty\else
        \footnotetext[1]{\AffiliationText \\  
        \ifx\EmailText\empty\else
            E-mail: \EmailText 
        \fi}
    \fi
}
\makeatother

\renewenvironment{abstract}{
    \begin{center}
        \textbf{Abstract}
    \end{center}
    \begin{quote}
}{
    \end{quote}
}

\titleformat{\section}
    {\normalfont\Large\bfseries}{\thesection.}{1em}{}

\titleformat{\subsection}
    {\normalfont\large\bfseries}{\thesubsection}{1em}{}

\title{Small Vision-Language Models: A Survey on Compact Architectures and Techniques}
\author{
    Nitesh Patnaik\textsuperscript{1}, 
    Navdeep Nayak\textsuperscript{1}, 
    Himani Bansal Agrawal\textsuperscript{1}, 
    Moinak Chinmoy Khamaru\textsuperscript{1}, 
    Gourav Bal\textsuperscript{1}, 
    Saishree Smaranika Panda\textsuperscript{1}, 
    Rishi Raj\textsuperscript{2}, 
    Vishal Meena\textsuperscript{1},
    Kartheek Vadlamani\textsuperscript{3}
}

\affiliation{
    \textsuperscript{1}School of Computer Engineering, Kalinga Institute of Industrial Technology (KIIT) Deemed to be University, Bhubaneswar-751024, Odisha, India.\\
    \textsuperscript{2}Information Systems, Indian Institute of Management Visakhapatnam, India.\\
    \textsuperscript{3}Remodel AI.
}

\email{
    \texttt{niteshpatnaik03@gmail.com, Rishi.raj@iimv.ac.in,
    kartheekvadlamani@gmail.com}
}

\begin{document}

\maketitle

\begin{abstract}
The emergence of small vision-language models (sVLMs) marks a critical advancement in multimodal AI, enabling efficient processing of visual and textual data in resource-constrained environments. This survey offers a comprehensive exploration of sVLM development, presenting a taxonomy of architectures—transformer-based, mamba-based, and hybrid—that highlight innovations in compact design and computational efficiency. Techniques such as knowledge distillation, lightweight attention mechanisms, and modality pre-fusion are discussed as enablers of high performance with reduced resource requirements. Through an in-depth analysis of models like TinyGPT-V, MiniGPT-4, and VL-Mamba, we identify trade-offs between accuracy, efficiency, and scalability. Persistent challenges, including data biases and generalization to complex tasks, are critically examined, with proposed pathways for addressing them. By consolidating advancements in sVLMs, this work underscores their transformative potential for accessible AI, setting a foundation for future research into efficient multimodal systems.
\end{abstract}

\begin{figure}[H]
                \centering
                \includegraphics[width=1\textwidth]{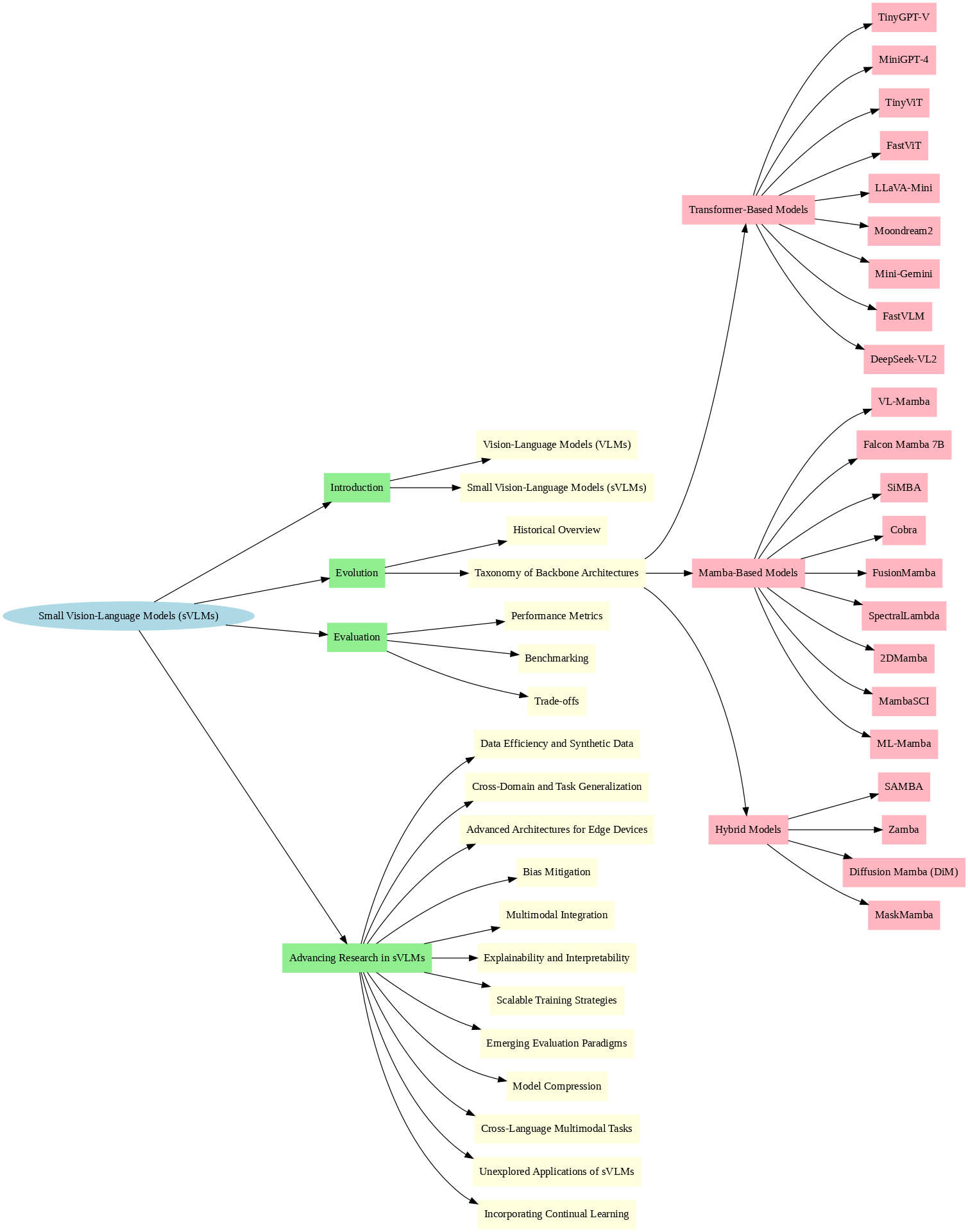}
                \caption{Structure of the Paper}
                \label{fig:sp}
\end{figure}

\section{Introduction} \label{section:intro}
Vision-language models (VLMs) have revolutionized multimodal artificial intelligence by enabling systems to understand and generate content from visual and textual data. These models find applications in various domains, including visual question answering (VQA) \cite{singh2019vqamodelsread}, image captioning, cross-modal retrieval, and document understanding. Early VLMs like CLIP (Contrastive Language–Image Pretraining) 
\cite{shen2021clipbenefitvisionandlanguagetasks} achieved groundbreaking performance by leveraging large-scale training datasets to learn generalized representations. However, these models are computationally intensive, requiring significant resources for both training and inference, limiting their applicability to resource-constrained environments. As the demand for efficient and scalable AI solutions grows, small vision-language models (sVLMs) have emerged as a critical area of research. These models aim to balance performance with computational efficiency, making advanced AI capabilities accessible for devices with limited resources, such as mobile phones, edge devices, and embedded systems. Unlike their larger counterparts, sVLMs are designed with compact architectures, leveraging innovative techniques like knowledge distillation, lightweight attention mechanisms, and modality pre-fusion to reduce computational overhead while retaining accuracy.  

Unlike other surveys that predominantly explore general VLMs or focus on large-scale models, this survey uniquely highlights compact architectures and techniques optimized for efficient use cases. It introduces a taxonomy of Samll Vision Language Models (sVLMs) categorized into transformer-based, Mamba-based, and hybrid models. We also critically evaluated these architectures' efficiency and scalability, providing a nuanced understanding of trade-offs between performance and resource usage. By consolidating advancements in techniques like knowledge distillation, lightweight attention mechanisms, and modality pre-fusion, it presents a focused lens on enabling multimodal AI accessibility across devices with limited computational capabilities. This paper aims to inform researchers and practitioners of the state-of-the-art advancements in sVLMs, identifying opportunities for future innovation and offering a roadmap for the development of efficient and versatile multimodal systems. By consolidating the progress in this rapidly evolving field, this survey serves as a resource for understanding the current landscape and guiding future research toward the next generation of vision-language models.

\begin{figure}[H]
                \centering
                \includegraphics[width=1\textwidth]{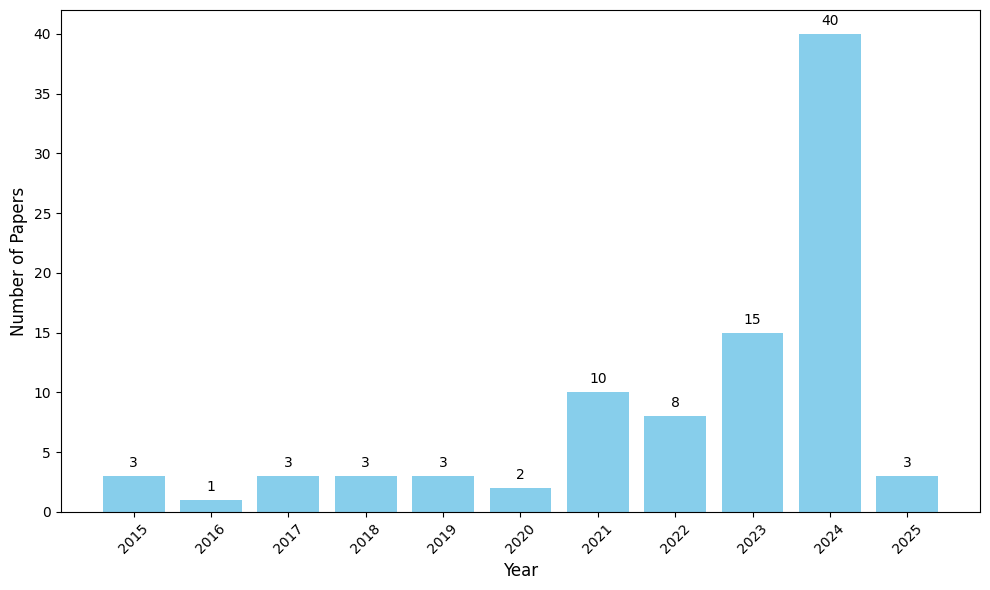}
                \caption{Number of Papers Published by Year}
                \label{fig:pp}
\end{figure}

\section{Evolution of Small Vision-Language Models} \label{sec:evolution}

\begin{figure}[H]
                \centering
                \includegraphics[width=1\textwidth]{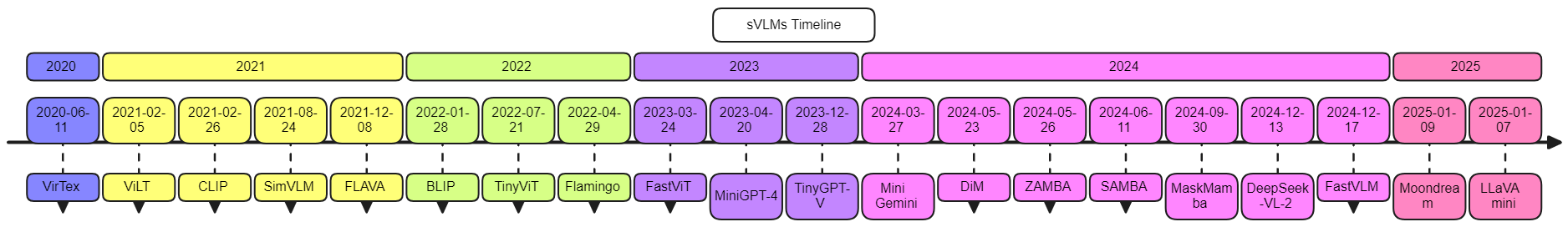}
                \caption{Evolution of Vision Language Models over time}
                \label{fig:timeline}
\end{figure}

\subsection{Historical Overview}

The development of small vision-language models (sVLMs) stems from the broader evolution of vision-language models (VLMs), which integrate visual and textual data to solve complex multimodal tasks. Early models like CLIP revolutionized the field by demonstrating the power of contrastive learning to align images and text in a shared embedding space, achieving remarkable zero-shot performance. However, the high computational demands of these models highlighted the need for more efficient alternatives. This section traces the historical trajectory of sVLMs, from foundational architectures like CLIP to emerging solutions that prioritize computational efficiency and scalability. It highlights the key breakthroughs and design principles that have shaped the evolution of sVLMs, setting the stage for the modern lightweight architectures discussed later in this survey.

\begin{figure}[H]
                \centering
                \includegraphics[width=1\textwidth]{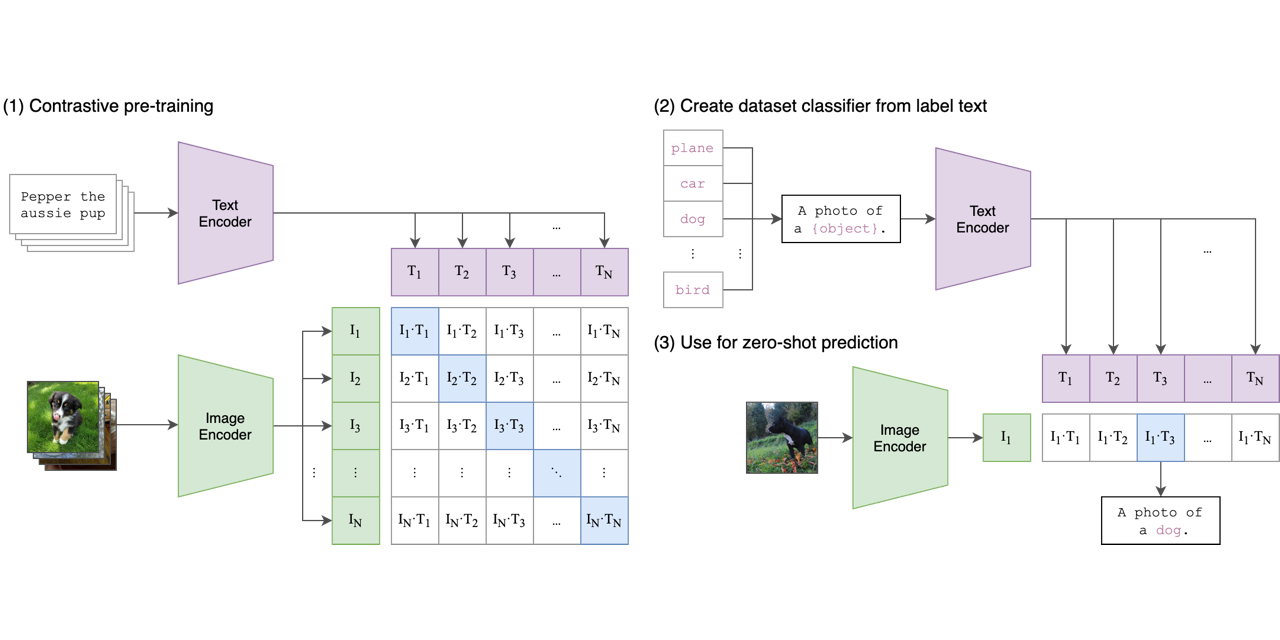}
                \caption{CLIP Architecture}
                \label{fig:clip}
\end{figure}

\textbf{Contrastive Language-Image Pre-training (CLIP)} \cite{radford2021learningtransferablevisualmodels} is an approach for training vision models through natural language supervision. Its architecture consists of an image encoder and a text encoder that project visual and textual inputs into a shared multi-modal embedding space. The components are trained using a contrastive learning objective that aligns correct image-text pairs while minimizing mismatched ones. CLIP demonstrates zero-shot performance across more than 30 datasets, predicting class labels for unseen tasks using natural language descriptions. However, its zero-shot performance remains below state-of-the-art models, requiring significant advancements in computational efficiency. CLIP struggles with fine-grained classification, abstract reasoning, and out-of-distribution generalization, particularly in handwritten digit recognition. It lacks flexibility in zero-shot classification, cannot generate novel outputs, and exhibits poor data efficiency, relying heavily on large-scale datasets. Furthermore, social biases in internet-sourced training data impact its fairness, and its few-shot performance unexpectedly deteriorates, necessitating further research in transfer learning optimization.

\textbf{ViLT (Vision-and-Language Transformer)} \cite{kim2021viltvisionandlanguagetransformerconvolution} introduces a Vision-and-Language Pre-training (VLP) approach that eliminates convolutional neural networks (CNNs) \cite{guo2022cmtconvolutionalneuralnetworks} and region-specific supervision, streamlining visual processing. It employs a transformer-only architecture where images are divided into 32×32 patches, linearly projected, and fused with textual embeddings for intra- and inter-modal interaction. Pre-trained on datasets like MSCOCO, Visual Genome, SBU Captions, and Google Conceptual Captions, ViLT demonstrates competitive performance across tasks such as Visual Question Answering (VQAv2) \cite{goyal2017makingvvqamatter}, Natural Language for Visual Reasoning (NLVR2) \cite{suhr-etal-2017-corpus}, and image-text retrieval. However, its simplified visual embedding lacks fine-grained object representation, leading to performance trade-offs in VQA due to the absence of explicit object detection. The reliance on patch projection limits detailed semantic understanding, affecting deep visual reasoning. Attempts to enhance representation through Masked Patch Prediction (MPP) \cite{wei2023maskedfeaturepredictionselfsupervised} proved ineffective, highlighting optimization challenges in visual masking. While efficient, scaling ViLT to larger models requires extensive computational resources and data. These limitations underscore the trade-offs in removing region-based embeddings, suggesting future improvements in fine-grained visual representations and enhanced pre-training strategies.

\textbf{VirTex} \cite{desai2021virtexlearningvisualrepresentations} introduces a pretraining approach that learns visual representations using semantically rich captions, providing a data-efficient alternative to traditional supervised and unsupervised methods reliant on large labeled datasets. The architecture consists of a ResNet-50 \cite{he2015deepresiduallearningimage} visual backbone and a bidirectional transformer-based textual head, jointly trained on image-caption pairs from the COCO Captions dataset \cite{lin2015microsoftcococommonobjects} to generate captions. After pretraining, the textual head is discarded, and the visual backbone is transferred to downstream tasks such as image classification, object detection, and instance segmentation. However, VirTex has several limitations. It relies on high-quality caption annotations, which limits scalability to large, noisy web datasets where such annotations are unavailable. Additionally, Masked Language Modeling (MLM) \cite{wei2023maskedfeaturepredictionselfsupervised} was tested but found to have poor sample efficiency and slower convergence, making it computationally expensive. The model also exhibits optimization challenges with larger visual backbones, as ResNet-101 did not consistently improve results over ResNet-50. Furthermore, fine-tuning is required for downstream tasks, adding computational overhead in resource-constrained environments. While VirTex excels in representation learning, its generated captions are suboptimal compared to state-of-the-art captioning models, highlighting a trade-off in performance for captioning tasks.

\textbf{FLAVA’s} \cite{singh2022flavafoundationallanguagevision} is a vision and language alignment model designed to handle vision, language, and multimodal tasks within a unified transformer-based architecture. It comprises an image encoder based on the ViT-B/16 architecture for processing visual inputs, a text encoder of the same design for textual representations, and a multimodal encoder that aligns and fuses information from both modalities. The model is pretrained on a dataset of 70 million image-text pairs from publicly available sources, ensuring accessibility and reproducibility. However, FLAVA has several limitations. The use of public datasets introduces inherent dataset biases that may impact generalization, despite efforts to enhance diversity. The smaller pretraining dataset, significantly smaller than those used by model - SimVLM (1.8B) \cite{wang2022simvlmsimplevisuallanguage}, limits scalability in large-scale applications. Additionally, optimization challenges arise due to the joint training of unimodal and multimodal tasks, involving complex objectives such as contrastive learning, masked multimodal modeling, and image-text matching. The harder learning process is further complicated by round-robin sampling, which lacks a structured curriculum for sequential task learning. Consequently, FLAVA may exhibit potential under-performance in vision-only tasks, as it is not as extensively trained on large-scale vision data.

\textbf{SimVLM} \cite{wang2022simvlmsimplevisuallanguage} introduces a simplified vision-language pretraining approach by employing a single prefix language modeling (PrefixLM) \cite{ding2024causallmoptimalincontextlearning} objective, which unifies bidirectional contextual representation and autoregressive text generation. Its architecture is based on a transformer encoder-decoder framework, where raw image patches are processed through a convolutional stage inspired by ResNet \cite{he2015deepresiduallearningimage}, and textual inputs are tokenized with positional encodings. The model is trained end-to-end on large-scale weakly supervised datasets, such as ALIGN \cite{tang2023alignsearchaligningads}, without relying on object detection modules or auxiliary losses. However, its reliance on noisy web-crawled data may result in suboptimal performance in tasks requiring fine-grained understanding. The absence of explicit region-level reasoning limits its effectiveness in capturing precise spatial relationships and object interactions. High computational requirements, including the use of TPU v3 chips, pose scalability challenges. Additionally, dependence on weak supervision constrains the model’s ability to learn detailed vision-language associations. In open-ended VQA \cite{Ishmam_2024} tasks, SimVLM struggles to generate meaningful responses without additional finetuning due to the noisiness of pretraining data. While demonstrating strong zero-shot capabilities, it still falls short in complex reasoning tasks compared to fully supervised models. Furthermore, its optimization for multimodal learning results in performance trade-offs in single-modality tasks like language understanding and image classification.

\begin{figure}[H]
                \centering
                \includegraphics[width=1\textwidth]{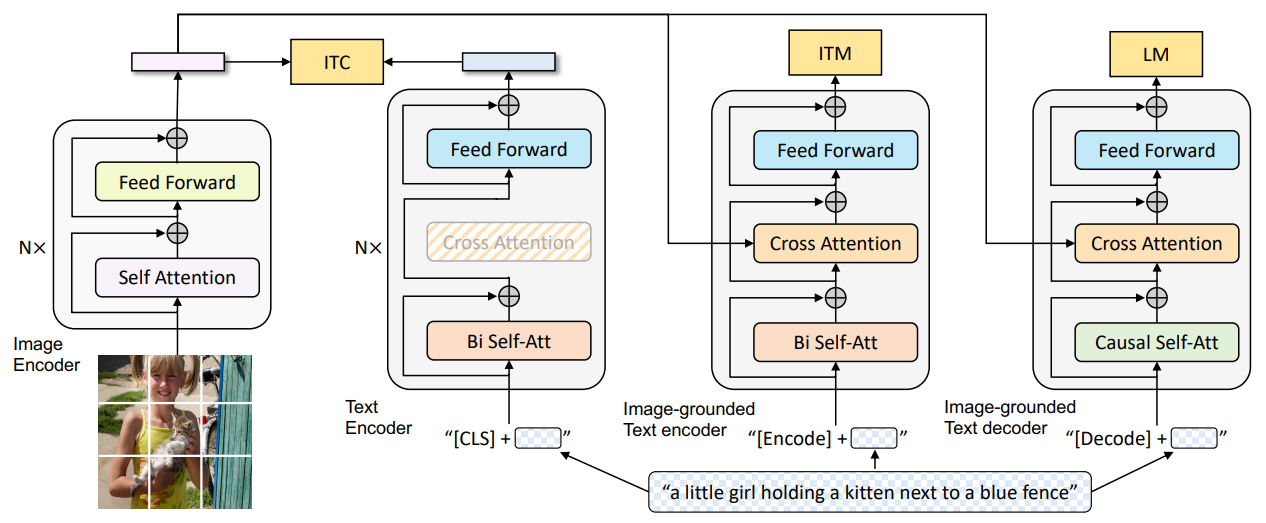}
                \caption{BLIP Architecture: Proposed a multimodal mixture
of encoder-decoder, a unified vision-language model which can operate in one of the three functionalities: (1) Unimodal encoder is
trained with an image-text contrastive (ITC) loss to align the vision and language representations. (2) Image-grounded text encoder uses
additional cross-attention layers to model vision-language interactions, and is trained with a image-text matching (ITM) loss to distinguish
between positive and negative image-text pairs. (3) Image-grounded text decoder replaces the bi-directional self-attention layers with
causal self-attention layers, and shares the same cross-attention layers and feed forward networks as the encoder. The decoder is trained
with a language modeling (LM) loss to generate captions given images.}
                \label{fig:blip}
\end{figure}

\textbf{BLIP (Bootstrapping Language-Image Pre-training)} \cite{li2022blipbootstrappinglanguageimagepretraining}, gives significant technical contributions to unified vision-language understanding and generation by employing a novel bootstrapped pre-training strategy that integrates text and image data efficiently. Its architecture consists of a vision encoder, based on the Vision Transformer (ViT) \cite{dosovitskiy2021imageworth16x16words}, and a text encoder-decoder, based on Transformer \cite{vaswani2023attentionneed}, working in tandem to align and generate multimodal representations. BLIP introduces two training objectives: image-grounded text generation and image-text contrastive learning, enabling robust pre-training on noisy web data. The model is evaluated on diverse tasks, including image captioning, visual question answering, and cross-modal retrieval, demonstrating state-of-the-art performance. It leverages large-scale datasets such as the Conceptual Captions \cite{changpinyo2021conceptual12mpushingwebscale} and MS COCO datasets \cite{lin2015microsoftcococommonobjects}, enriched with web-crawled image-text pairs for pre-training. The limitations of BLIP stem from both model and data perspectives. On the model side, encoder-based approaches struggle to directly apply to text generation tasks, such as image captioning, while encoder-decoder models face challenges in excelling at understanding-based tasks like image-text retrieval. From a data perspective, BLIP, like other vision-language pre-training methods, heavily relies on noisy web data, which is suboptimal for vision-language learning. These noisy web texts often fail to accurately describe the visual content of images, leading to inefficiencies in learning robust vision-language alignment despite gains from scaling up datasets.

\begin{figure}[H]
                \centering
                \includegraphics[width=1\textwidth]{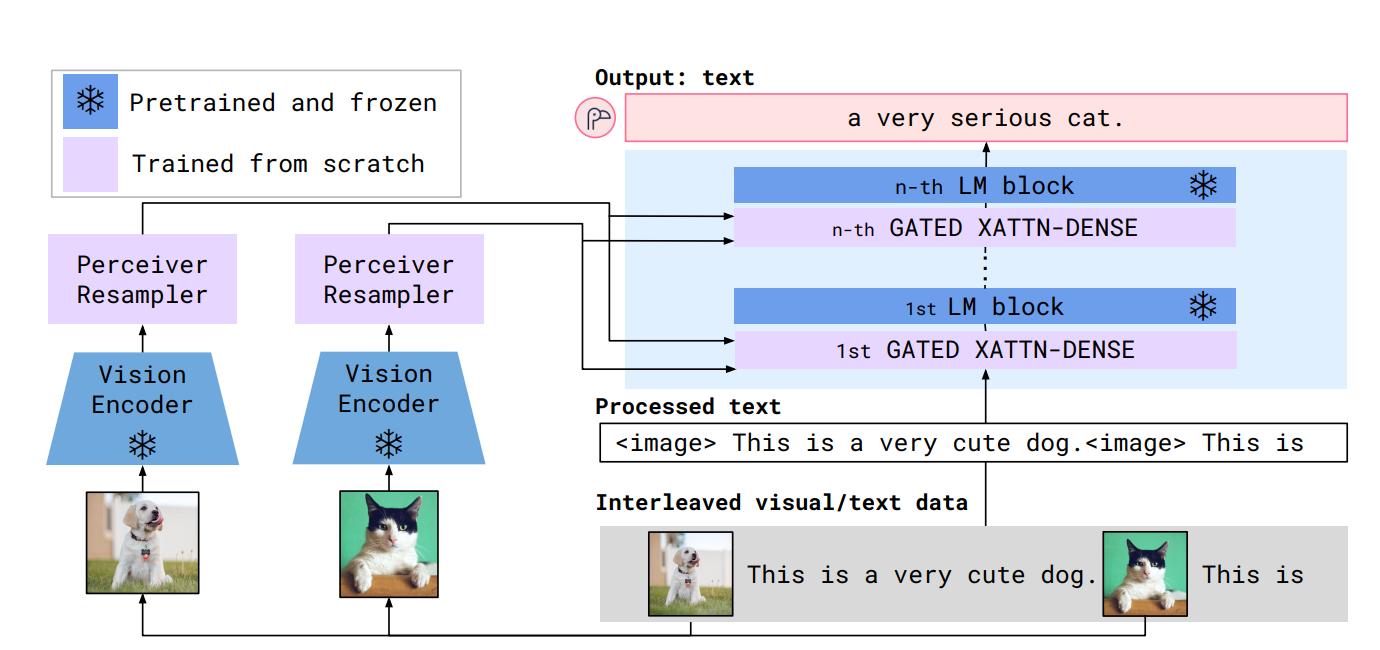}
                \caption{Flamingo Architecture: Takes input visual data interleaved with text and produce free-form text as output. Introduction of GATED XATTN-DENSE layers to condition the Language Model on visual inputs}
                \label{fig:flamingo}
\end{figure}

\textbf{Flamingo} \cite{alayrac2022flamingovisuallanguagemodel} is a visual language model that bridges pretrained vision-only and language-only models to handle sequences of interleaved visual and textual data, enabling state-of-the-art few-shot learning on multimodal tasks \cite{lin2024multimodalityhelpsunimodalitycrossmodal}. Its architecture integrates a frozen NFNet-F6 \cite{brock2021highperformancelargescaleimagerecognition} vision encoder with a Perceiver Resampler \cite{jaegle2021perceivergeneralperceptioniterative} to convert variable-sized visual features into fixed tokens, which are processed by a frozen large language model augmented with newly added GATED XATTN-DENSE layers. These innovations allow efficient cross-modal integration while preserving the pretrained language model’s capabilities. Flamingo was trained on a diverse mixture of datasets, including the M3W dataset \cite{zhu2023multimodal} (43 million webpages with interleaved text and images), ALIGN (1.8 billion image-text pairs) \cite{jia2021scalingvisualvisionlanguagerepresentation} and LTIP (312 million long-text and image pairs) \cite{evans2024badstudentsmakegreat}, using a novel multi-objective loss function. The model was evaluated on 16 multimodal benchmarks, achieving new state-of-the-art results in few-shot learning while using significantly less task-specific data than previous models. Despite its strengths, Flamingo, like other models built on pretrained language models (LMs), inherits several inherent weaknesses from its foundation. For instance, while LM priors are often beneficial, they can occasionally lead to hallucinations or ungrounded predictions. Additionally, LMs demonstrate poor generalization to sequences longer than those observed during training and suffer from low sample efficiency during the training process. Addressing these limitations could significantly enhance the effectiveness and reliability of vision-language models (VLMs) such as Flamingo.Another limitation of Flamingo lies in its classification performance, which falls short of state-of-the-art contrastive models. Contrastive models excel at tasks like text-image retrieval by directly optimizing for such objectives, with classification as a subset. In contrast, Flamingo is designed to address a wider range of open-ended tasks. Developing unified approaches that combine the strengths of contrastive models and the versatility of Flamingo represents an important direction for future research.In-context learning, one of Flamingo's key strengths, also presents challenges. This approach allows for simple deployment and inference with minimal hyperparameter tuning and has demonstrated effectiveness in low-data scenarios, particularly with only a few dozen examples. However, its sensitivity to demonstration setups, high inference compute costs, and diminishing performance as the number of examples scales beyond low-data regimes are notable drawbacks. Combining in-context learning with gradient-based few-shot methods could provide a more balanced and effective solution.

\subsection{Taxonomy of backbone Architecture in small Vision-Language Models (sVLMs)}

The backbone architecture is pivotal to the design of small vision-language models (sVLMs), serving as the foundation for integrating visual and textual modalities. While existing research often focuses on individual model innovations, our survey introduces a novel classification system that categorizes these architectures into three distinct paradigms based on their structural and functional principles:  

\begin{enumerate}  
    \item \textbf{Transformer-Based Models \cite{vaswani2023attentionneed}:} These models dominate the landscape with their powerful self-attention mechanisms, enabling nuanced multimodal understanding and efficient representation learning.  
    \item \textbf{Mamba-Based Models \cite{gu2024mambalineartimesequencemodeling}:} Offering a departure from transformer-centric designs, these architectures leverage state-space models for linear scalability and computational efficiency, making them highly suitable for handling long sequences and resource-constrained scenarios.  
    \item \textbf{Hybrid Models:} These architectures combine elements from transformers, convolutional neural networks (CNNs) \cite{guo2022cmtconvolutionalneuralnetworks}, and other lightweight mechanisms to create versatile systems that balance task performance with computational demands.  
\end{enumerate}  

This classification provides a cohesive framework for understanding the diverse strategies employed in the development of sVLMs. By organizing models into these categories, we highlight not only their unique design principles but also the trade-offs and advantages they offer across various applications. This section emphasizes the importance of this taxonomy in advancing the field and serves as a foundation for exploring the state-of-the-art innovations in sVLMs.

\subsubsection{Transformer-Based Models}

\begin{figure}[H]
                \centering
                \includegraphics[width=1\textwidth]{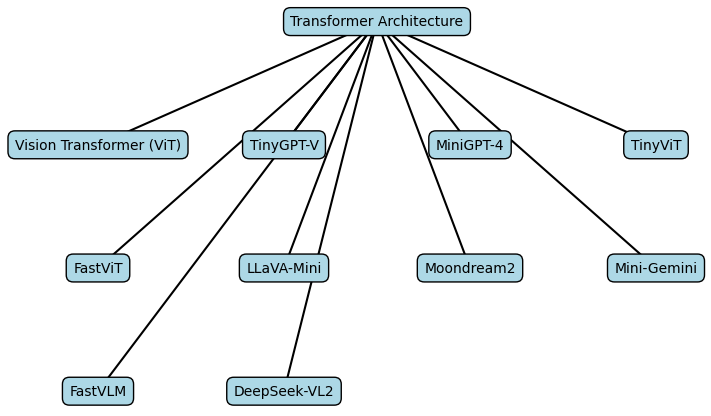}
                \caption{Models based on Transformer Architecture}
                \label{fig:transmod}
\end{figure}

The transformer architecture \cite{vaswani2023attentionneed} is built  on a self-attention mechanism that enables the model to weigh the importance of different elements in a sequence, allowing it to capture long-range dependencies and complex contextual relationships. This architecture, initially developed for natural language processing tasks, has been successfully adapted for small vision-language models (sVLMs), where it efficiently handles both visual and textual modalities. In transformer-based sVLMs, the two modalities—images and text—are processed in parallel, each modality being encoded separately before being fused for multimodal tasks. The image input is typically divided into fixed-size patches, which are then linearly embedded into vectors, akin to the tokenization process used for text. These visual tokens are then passed through a series of transformer layers, where the self-attention mechanism enables the model to capture the relationships between different image patches. For the textual modality, the input text is tokenized into subword units, and the resulting embeddings are processed similarly through the transformer layers. The fusion of these modalities typically occurs in one of two ways: early fusion or late fusion. A key innovation in these models is the use of cross-attention mechanisms, where the textual features attend to the visual features and vice versa. This enables the model to focus on the most relevant parts of each modality when performing tasks such as visual question answering (VQA) or image captioning. By allowing the model to learn both intra-modal (within the same modality) and inter-modal (between modalities) relationships, transformers can effectively handle complex multimodal data. The transformer architecture in sVLMs provides a powerful and flexible framework for integrating visual and textual information. Its ability to model long-range dependencies, capture contextual relationships, and fuse multimodal data efficiently makes it an ideal choice for vision-language tasks.

\begin{figure}[H]
                \centering
                \includegraphics[width=1\textwidth]{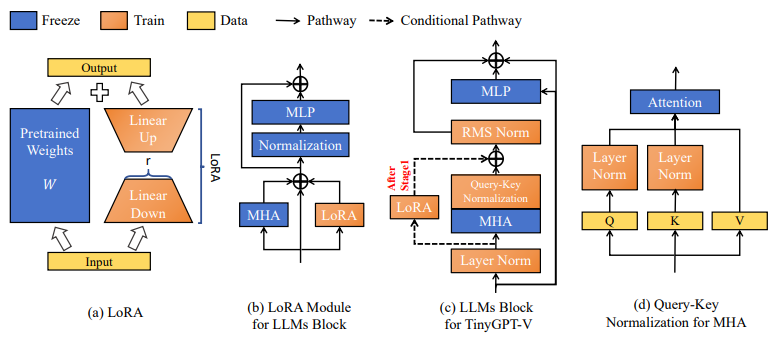}
                \caption{TinyGPT-V Architecture:  Highlighting key components such as the Conditional Pathway, Layer Normalization, and LoRA (Low-Rank Adaptation) modules for efficient adaptation in large language models (LLMs). It also depicts the integration of Multi-Head Attention (MHA) mechanisms and normalization layers within the LLMs block, emphasizing the streamlined design for enhanced performance in conditional tasks.}
                \label{fig:tinygpt}
\end{figure}

\textbf{TinyGPT-V} \cite{yuan2024tinygptvefficientmultimodallarge} introduces several key technical contributions, including an efficient architecture that integrates the compact Phi-2 \cite{Microsoft_Phi_2} language model with pre-trained vision encoders, a novel training methodology focused on small pre-trained backbones, and resource-efficient training and inference capabilities. The model addresses challenges such as the closed-source nature and high computational demands of large MLLMs, aiming to achieve comparable performance to larger models while using significantly fewer resources. TinyGPT-V's architecture consists of a visual encoder using EVA (ViT) \cite{fang2022evaexploringlimitsmasked}, projection layers including Q-Former layers and linear projections, and the Phi-2 language model as the backbone, incorporating various normalization techniques and LoRA \cite{hu2021loralowrankadaptationlarge} for efficient fine-tuning. The model is evaluated on benchmarks such as GQA Hudson \cite{hudson2019gqanewdatasetrealworld}, VSR \cite{zhang2021vsrunifiedframeworkdocument}, IconVQ \cite{lu2021iconqa}, VizWiz \cite{8954403}, and Hateful Memes (HM) \cite{kiela2021hatefulmemeschallengedetecting}, using a diverse set of datasets across its training stages. The ablation study highlights the potential challenges in training smaller language models for multimodal tasks, which is one of the possible drawbacks. The removal of key components like LoRA, Input Layer Norm, RMS Norm, and QK Norm leads to significant training issues, including gradient vanishing and increased loss. TinyGPT-V's performance relies heavily on specific normalization techniques and architectural components for stable training of smaller models, which could be considered a limitation of the approach. Additionally, there may be trade-offs between model size and performance on certain tasks, as evidenced by the comparative results with larger models.

\textbf{MiniGPT-4} \cite{zhu2023minigpt4enhancingvisionlanguageunderstanding} presents a novel approach to enhancing vision-language understanding by aligning a frozen visual encoder with an advanced large language model (LLM), specifically Vicuna \cite{vicuna2023}. The model addresses the challenge of replicating GPT-4's impressive multi-modal abilities while keeping the technical details transparent. MiniGPT-4's architecture consists of a vision encoder with a pretrained ViT and Q-Former, a single linear projection layer, and the Vicuna LLM. Only the linear projection layer is trained to align visual features with Vicuna, while all other components remain frozen. The model undergoes a two-stage training process. In the first stage, it is pretrained on a combined dataset of approximately 5 million image-text pairs from LAION \cite{schuhmann2022laion5bopenlargescaledataset} , Conceptual Captions \cite{changpinyo2021conceptual12mpushingwebscale} and SBU \cite{NIPS2011_5dd9db5e}. The second stage involves fine-tuning with a curated dataset of 3,500 detailed image-text pairs to improve generation reliability and usability. Evaluation metrics include performance on traditional visual question answering (VQA) benchmarks such as A-OKVQA and GQA \cite{hudson2019gqanewdatasetrealworld}, as well as qualitative assessments of various multi-modal tasks. MiniGPT-4 demonstrates several advanced capabilities, including detailed image description generation, website creation from hand-drawn drafts, meme interpretation, and creative tasks like poem writing and recipe generation. However, the model has limitations, such as occasional hallucinations in image descriptions and challenges in understanding spatial information. The study also reveals that simply aligning visual features with LLMs using short image caption pairs is insufficient for developing a well-performing model, highlighting the importance of the second-stage fine-tuning process.

\begin{figure}[H]
                \centering
                \includegraphics[width=1\textwidth]{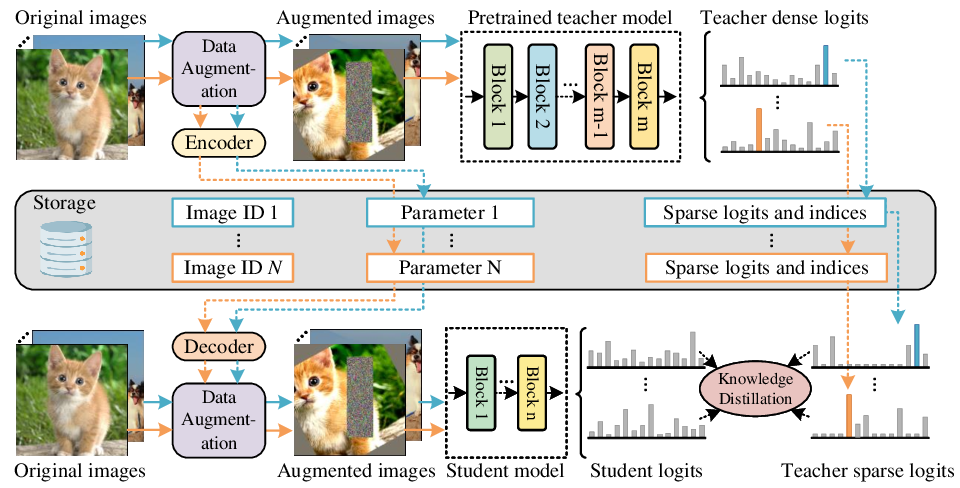}
                \caption{TinyViT Architecture: A pretrained teacher model generates dense logits from original and augmented images. These logits are sparsified and stored, then used to guide the training of a student model. The student model, through a series of encoder and decoder blocks, learns to replicate the teacher's sparse logits, enhancing efficiency and performance through knowledge distillation and data augmentation techniques.}
                \label{fig:tinyvit}
\end{figure}

\textbf{TinyViT} \cite{wu2022tinyvitfastpretrainingdistillation} is a group of efficient vision transformers(ViT) \cite{yuan2021tokenstotokenvittrainingvision} designed for devices with limited computing power, like mobile phones and edge devices. It employs knowledge distillation, where it learns from 11 larger, more powerful models to improve its performance. TinyViT solves the problem of high memory and processing demands in traditional vision transformers, making them practical for real-time use. The architecture features a hierarchical structure with progressive downsampling and localized attention to balance efficiency and accuracy. It is tested on popular benchmarks like ImageNet \cite{russakovsky2015imagenetlargescalevisual} for image classification, COCO \cite{lin2015microsoftcococommonobjects} for object detection, and ADE20K \cite{zhou2018semanticunderstandingscenesade20k} for semantic segmentation. However, large model sizes and the heavy computational costs make these models unsuitable for applications with real-time requirements. In the image classification field, lightweight networks frequently incorporate a hierarchical architecture with high-stride downsampling to decrease computational expenses. Large-stride downsampling often leads to a loss of critical information, which is crucial for accurate tracking.

\textbf{FastViT} \cite{vasu2023fastvitfasthybridvision} is a hybrid vision transformer designed for good speed and accuracy, offering a state-of-the-art balance between latency and performance. Its key innovation is the RepMixer \cite{bui2022repmixrepresentationmixingrobust} block, which eliminates skip connections through structural reparameterization, reducing memory access costs and boosting efficiency. The architecture also integrates train-time overparameterization and large kernel convolutions to enhance accuracy without increasing latency. FastViT addresses the inefficiencies of traditional self-attention and skip connections, outperforming models like CMT \cite{guo2022cmtconvolutionalneuralnetworks}, EfficientNet \cite{masters2021makingefficientnetefficientexploring}, and ConvNeXt \cite{10351320} on tasks such as image classification, object detection, and semantic segmentation. It is evaluated on datasets like ImageNet-1K \cite{cui2023scalingdatasetdistillationimagenet1k}, MS COCO \cite{zimmermann2023benchmarkingbenchmarkreliablemscoco}, and ADE20K \cite{zhou2018semanticunderstandingscenesade20k}. However, train-time overparameterization results in increased training time due to computational overhead. Its robustness to corruptions and out-of-distribution samples is notable but not consistent. The model’s reliance on hardware-specific optimizations, like Core ML for mobile devices, can limit its adaptability to other platforms.

\textbf{LLaVA-Mini} \cite{zhang2025llavaminiefficientimagevideo} is an efficient large multimodal model (LMM) \cite{tzelepi2024lmmregularizedclipembeddingsimage} designed to reduce computational costs while maintaining strong visual understanding capabilities. Its key innovation is the modality pre-fusion module, which integrates visual information into text tokens before processing, allowing the number of vision tokens per image to be compressed from 576 to just 1. This approach drastically reduces FLOPs, memory usage, and latency, making it ideal for high-resolution image and video tasks. The architecture incorporates a query-based compression module alongside pre-trained CLIP \cite{shen2021clipbenefitvisionandlanguagetasks} vision encoders and Vicuna language models, enabling efficient processing without sacrificing accuracy. Evaluated across 11 image-based and 7 video-based benchmarks, including datasets like ImageNet \cite{anzaku2024reassessingimagenetalignedsinglelabel}, COCO \cite{lin2015microsoftcococommonobjects}, and MSVD-QA \cite{xu2017video}, LLaVA-Mini delivers better performance than previous models. The limitations include, the extreme compression of vision tokens into a single token risks losing visual details, particularly in high-resolution images. Its performance relies heavily on query-based compression and modality pre-fusion, with noticeable degradation if these strategies are absent. Although it handles long videos well through extrapolation, its training is limited to short videos, which could affect robustness in real-world scenarios.

\textbf{Moondream2} \cite{moondream2} is a cutting-edge multimodal system designed to handle image and video tasks efficiently through a hybrid architecture that combines modality pre-fusion and query-based compression. Its key innovation lies in compressing vision tokens into compact representations while retaining semantic richness, reducing computational and memory requirements. This design addresses the high resource demands of traditional multimodal models, particularly in processing large-scale visual data. By integrating compressed vision tokens with text tokens before feeding them into the backbone model, Moondream2 achieves exceptional efficiency without compromising performance. It is evaluated on diverse benchmarks, such as MVBench \cite{li2024mvbenchcomprehensivemultimodalvideo} and MLVU \cite{zhou2025mlvubenchmarkingmultitasklong}. The datasets include both image-based and video-based tasks, covering scenarios like visual question answering and long-form video comprehension. Moondream2, while efficient for edge devices, has limitations including generating inaccurate statements, struggling with nuanced instructions, and exhibiting potential biases. It may also produce offensive or inappropriate content if not carefully guided. It relies on extensive pretraining and may face challenges in capturing fine-grained visual details or handling tasks that require richer global context.

\textbf{Mini-Gemini} \cite{li2024minigeminiminingpotentialmultimodality} is a cutting-edge vision-language framework that enhances multimodal
vision-language models (VLMs) \cite{xu2021vlmtaskagnosticvideolanguagemodel} by refining visual tokens, improving dataset quality, and
advancing reasoning capabilities. Its key innovation is a dual-encoder system that processes high-resolution and low-resolution inputs, enabling detailed visual comprehension through patch info mining while maintaining efficiency by keeping the token count fixed. Mini-Gemini addresses challenges like computational inefficiency in handling high-resolution images and enhances vision-language alignment through a curated dataset of over 2.7 million high-quality captions and instructions. The model supports simultaneous text and image understanding and generation, achieving state-of-the-art results across benchmarks such as VQAT \cite{singh2019vqamodelsread}, MMB \cite{liu2024mmbenchmultimodalmodelallaround}, and MM-Vet \cite{yu2024mmvetevaluatinglargemultimodal}. Mini-Gemini faces limitations in counting and complex visual reasoning tasks, largely due to insufficient specialized training data during pretraining. Its reasoning-based generation relies on text to bridge the Vision-Language Model and diffusion model, as
embedding-based methods have not shown significant improvements.

\begin{figure}[H]
                \centering
                \includegraphics[width=1\textwidth]{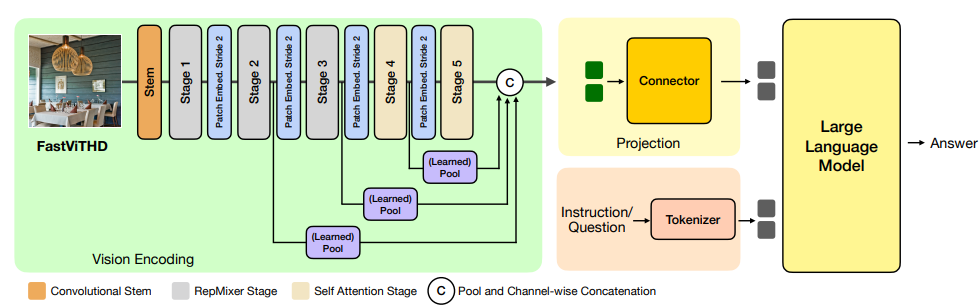}
                \caption{FastVLM Architecture: Features a bi-convolutional stem for initial feature extraction, followed by a RepMixer stage that incorporates self-attention mechanisms. The process concludes with pooling and channel-wise concatenation to integrate and refine visual features for efficient vision-language modeling.
}
                \label{fig:fastvlm}
\end{figure}

\textbf{FastVLM} \cite{vasu2023fastvitfasthybridvision} introduces a vision-language architecture featuring the FastViTHD encoder, a hybrid convolution-transformer design optimized for high-resolution input. By significantly reducing encoding latency and visual token count, it achieves a good accuracy-latency trade-off without relying on complex techniques like token pruning or dynamic resolution. Leveraging multi-scale features and scalable input resolution, FastVLM demonstrates state-of-the-art performance on benchmarks such as SeedBench \cite{li2023seedbenchbenchmarkingmultimodalllms} and MMMU \cite{yue2024mmmumassivemultidisciplinemultimodal}, particularly excelling in text-rich and general knowledge tasks. Trained on datasets like DataCompDR-1B \cite{gadre2023datacompsearchgenerationmultimodal} and LLaVA-558K \cite{xu2017video,liu2023improvedllava,liu2023llava}, it ensures scalability and efficiency.However, FastVLM has several limitations, At very high image resolutions (i.e 1536x1536), performance is constrained by on-device memory bandwidth. While dynamic resolution (tiling) is efficient at high resolutions, it is less optimal compared to single-resolution input when memory isn't a constraint. FastVLM relies heavily on large training datasets for optimal performance and faces challenges in resource-constrained environments due to its vision encoder size, which is 3.5× larger than the largest MobileCLIP-based FastViT variant.

\textbf{DeepSeek-VL2} \cite{wu2024deepseekvl2mixtureofexpertsvisionlanguagemodels} is a Vision-Language Model (VLM) \cite{bordes2024introductionvisionlanguagemodeling} architecture with a dynamic tiling vision encoding strategy that efficiently handles high-resolution images with varying aspect ratios, addressing the limitations of fixed-resolution encoders in tasks like visual grounding and OCR \cite{liu2024tellenhancingkeyinformation}. Leveraging a Mixture-of-Experts (MoE) \cite{cai2024surveymixtureexperts} framework, the model uses sparse computation to scale efficiently across three variants (1.0B, 2.8B, and 4.5B activated parameters), while the Multi-head Latent Attention (MLA) mechanism compresses Key-Value caches into latent vectors, optimizing inference speed. Trained on a high-quality, diverse vision-language dataset, DeepSeek-VL2 excels in tasks such as document understanding, chart interpretation, and visual grounding.However,It currently has a limited context window for images, restricting its ability to handle complex multi-image interactions like reasoning or understanding relationships across multiple images. Additionally, it struggles with blurry images and unseen objects, requiring enhanced robustness to manage diverse visual inputs and variations in quality. While the model excels in visual perception and recognition, its reasoning capabilities need further development to perform more complex reasoning tasks based on visual input.

\subsubsection{Mamba-Based Models}

\begin{figure}[H]
                \centering
                \includegraphics[width=1\textwidth]{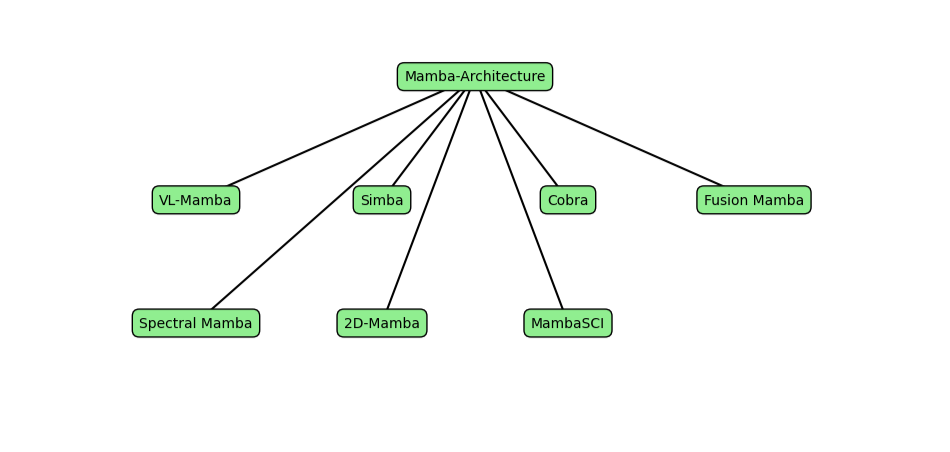}
                \caption{Models based on Mamba Architecture}
                \label{fig:mambamod}
\end{figure}

\begin{figure}[H]
                \centering
                \includegraphics[width=1\textwidth]{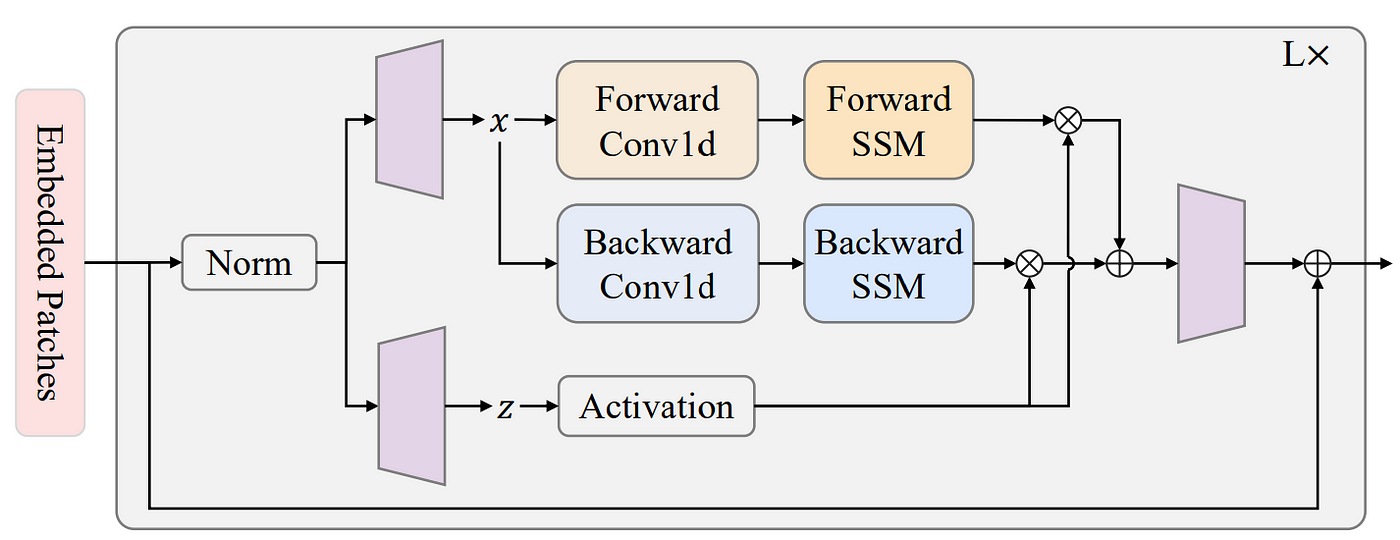}
                \caption{Vision Mamba (ViM) architecture: Showcases the processing of embedded patches through a series of operations. These include normalization, 1D convolutions, and state space models (SSM) with forward and backward passes. The encoder leverages activation functions and structured state management to efficiently process visual data, highlighting the integration of advanced sequence modeling techniques for vision tasks.}
                \label{fig:vim}
\end{figure}

Mamba-based \cite{gu2024mambalineartimesequencemodeling} models represent a significant innovation in lightweight backend architectures, specifically designed for efficiency and scalability in various machine learning applications. Mamba leverages the Structured State Space (S4) model to effectively manage long data sequences, addressing the limitations inherent in traditional transformer models, particularly their inefficiencies when processing lengthy inputs. By integrating continuous-time, recurrent, and convolutional models, Mamba adeptly handles irregularly sampled data and unbounded contexts while maintaining computational efficiency during both training and inference. Key feature of Mamba's architecture is its selective state space mechanism, which dynamically adjusts its parameters based on the input data. This adaptability allows the model to focus on relevant information within sequences and filter out extraneous data, thereby enhancing its ability to prioritize significant data points for various tasks and improving overall performance. Mamba simplifies the architecture by replacing complex attention mechanisms and multi-layer perceptron (MLP) blocks with a unified state space model block. This design choice not only reduces computational complexity but also accelerates inference speeds, making Mamba particularly well-suited for real-time applications.The performance of Mamba is further augmented by hardware-aware algorithms that exploit GPU capabilities through techniques such as kernel fusion and parallel scanning. These optimizations minimize memory usage and enhance performance when processing long sequences, achieving an impressive balance between resource utilization and computational power.

\begin{figure}[H]
                \centering
                \includegraphics[width=1\textwidth]{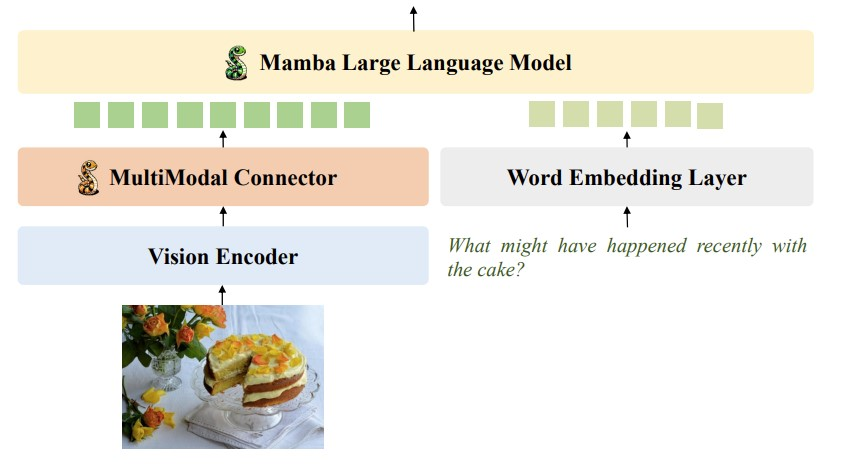}
                \caption{VL-Mamba Architecture: contains a Vision Encoder, a MultiModal Connector
(MMC), and a language model.}
                \label{fig:vlmamba}
\end{figure}

\textbf{VL-Mamba} \cite{qiao2024vlmambaexploringstatespace} presents several technical contributions, primarily centered on introducing one of the first multimodal large language model (MLLM) based on state space models (SSMs) rather than conventional Transformer architectures. VL-Mamba incorporates the Mamba large language model, which offers efficient long-sequence modeling with linear scalability and fast inference, replacing the computationally intensive attention mechanisms of Transformers. A key innovation is the MultiModal Connector (MMC) \cite{lin2024preservecompressindepthstudy}, which integrates a Vision Selective Scan (VSS) \cite{huang2024localmambavisualstatespace} module to bridge the gap between 2D non-causal image data and 1D causal sequence modeling in SSMs. Two 2D scanning mechanisms, the Bidirectional-Scan Mechanism (BSM) and the Cross-Scan Mechanism (CSM), are proposed to capture diverse spatial contexts efficiently. The MMC is available in three architectural variants—MLP, VSS-MLP, and VSS-L2—with VSS-L2 demonstrating superior performance in capturing and aggregating multimodal features. The vision encoder leverages pre-trained Vision Transformers to extract image patch features, which are processed by the MMC and combined with textual embeddings from the Mamba LLM for coherent vision-language reasoning and generation. Extensive experiments conducted on benchmarks such as VQA-v2 \cite{agrawal2016vqavisualquestionanswering}, GQA \cite{hudson2019gqanewdatasetrealworld}, and ScienceQA-IMG \cite{lu2022learnexplainmultimodalreasoning} evaluate the model's ability to perform tasks like visual question answering and multimodal reasoning. The reliance on data quality for further performance improvement is noted as a limitation, indicating opportunities for future work.

\begin{figure}[H]
                \centering
                \includegraphics[width=1\textwidth]{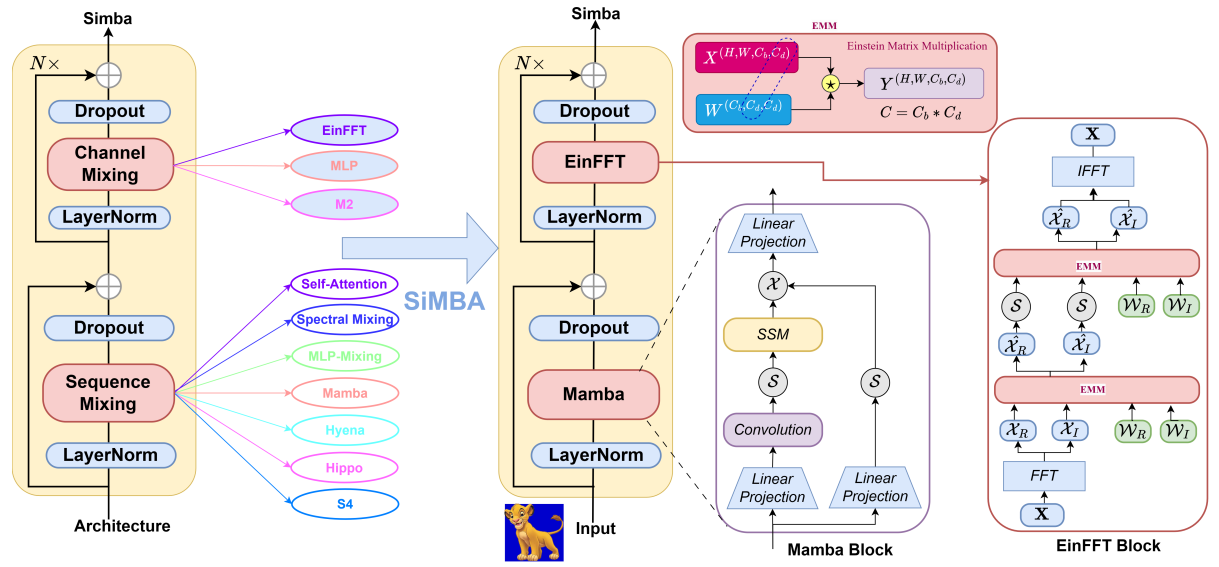}
                \caption{SiMBA Architecture: Features components like channel mixing, layer normalization, self-attention, spectral mixing, and MLP-mixing. The architecture integrates state space models (SSM), Mamba, and Hyena mechanisms, along with Fast Fourier Transform (FFT) operations for efficient feature extraction and transformation. Simba is designed to handle complex visual inputs through a series of linear projections, convolutions, and embedding matrix multiplications, ensuring robust and efficient performance in vision tasks.}
                \label{fig:simba}
\end{figure}

\textbf{SiMBA} \cite{patro2024simbasimplifiedmambabasedarchitecture}, introduces EinFFT, a novel channel modeling technique utilizing Fourier transforms to stabilize Mamba models at larger scales. It combines Mamba for sequence modeling and EinFFT for channel mixing, effectively addressing Mamba's instability and surpassing existing State Space Models (SSMs). This architecture also incorporates residual connections and dropouts to improve stability and performance. It closes the performance gap with state-of-the-art transformers on ImageNet \cite{russakovsky2015imagenetlargescalevisual} and time series datasets using ImageNet-1K \cite{cui2023scalingdatasetdistillationimagenet1k}, multivariate time series datasets, CIFAR-10/100 \cite{sun2024cifar10warehousebroadrealistictestbeds}, Stanford Cars \cite{6755945}, MS COCO \cite{lin2015microsoftcococommonobjects}, and ADE20K \cite{zhou2018semanticunderstandingscenesade20k} for evaluation. However, it still faces limitations including: a performance gap for large networks compared to top-tier transformers, instability issues with the MLP channel mixing variant at large scales, and remaining instability issues inherent to the Mamba architecture itself. There are also performance trade-offs with different channel mixing methods, and ablation studies show a performance gap when S4 and Hippo are used as standalone models without channel mixing. This suggests that its performance is dependent on the combination of sequence and channel mixing components it employs, indicating opportunities for further research.

\begin{figure}[H]
                \centering
                \includegraphics[width=1\textwidth]{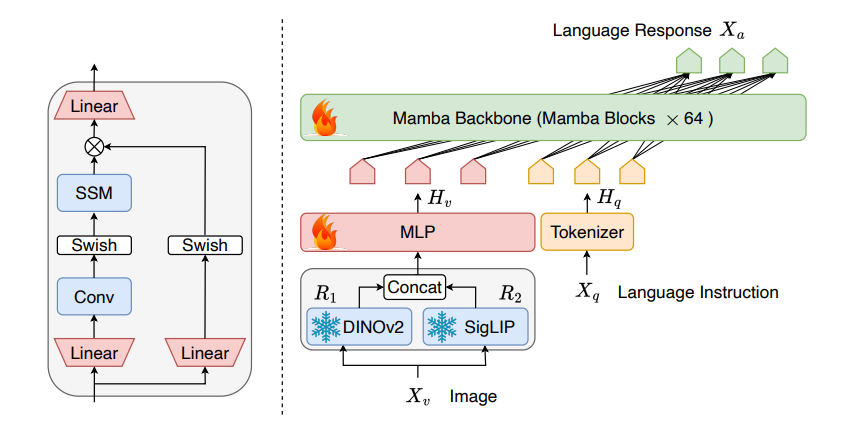}
                \caption{Cobra Architecture: Integrates a Mamba backbone with 64 Mamba blocks for processing language and visual inputs. It combines linear transformations, state space models (SSM), Swish activations, and convolutional layers to handle complex data. The architecture uses tokenizers and MLPs to process language instructions and image inputs ($X_v$) from models like DINOv2 and SigLIP, concatenating responses ($R_1$ and $R_2$) to generate coherent language outputs ($X_a$). This design ensures efficient and effective multimodal processing for tasks involving both text and images.}
                \label{fig:cobra}
\end{figure}

\textbf{Cobra} \cite{zhao2025cobraextendingmambamultimodal}, it employs a vision encoder that fuses features from DINOv2 \cite{oquab2024dinov2learningrobustvisual}, which captures low-level spatial details, and SigLIP \cite{zhai2023sigmoidlosslanguageimage}, which provides semantic visual representations. These features are combined into a compact representation. A projector, implemented as a learnable module, aligns the visual and textual modalities by mapping visual features into the language embedding space. Cobra’s backbone is the Mamba language model, which utilizes a stack of SSM-based layers, including convolutional components, selective propagation mechanisms, residual connections, and Root Mean Square Normalization (RMSNorm). The architecture processes input sequences auto-regressively, effectively integrating visual and text tokens to generate natural language responses. Key contributions include Cobra's ability to achieve inference speeds 3× to 4× faster than leading models like LLaVA-Phi \cite{zhu2024llavaphiefficientmultimodalassistant} and MobileVLM \cite{chu2023mobilevlmfaststrong}, while maintaining or exceeding performance on various benchmarks. It reduces parameter count by 48\% without sacrificing accuracy, demonstrating efficiency on both open-ended tasks (e.g., VQA-v2, TextVQA \cite{singh2019vqamodelsread}) and closed-set tasks (e.g., VSR \cite{zhang2021vsrunifiedframeworkdocument}, POPE \cite{li2023evaluatingobjecthallucinationlarge}). Limitations include a dependence on vision encoder choice, the negative impact of lightweight down-sample projectors (LDPv2), the importance of training strategies including the number of training epochs and projector initialization, the influence of model parameter size on the grounding ability, and sensitivity to prompt order, particularly when using OCR tokens.

\begin{figure}[H]
                \centering
                \includegraphics[width=1\textwidth]{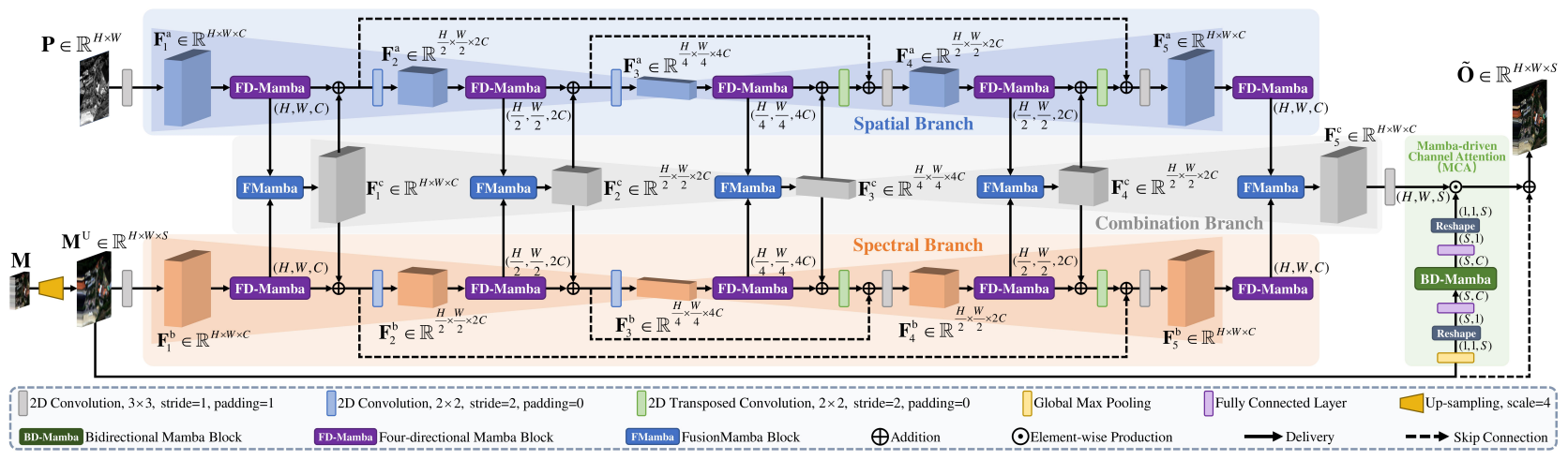}
                \caption{FusionMamba Architecture:  Incorporates channel attention mechanisms and multi-channel attention (MCA) to enhance feature representation. The architecture includes 2D convolutions, transposed convolutions, max pooling, and fully connected layers for efficient feature extraction and upscaling. FusionMamba is designed to handle complex data transformations, ensuring robust performance in tasks requiring detailed feature integration and spatial processing.}
                \label{fig:fm}
\end{figure}

\textbf{FusionMamba} \cite{Peng_2024} presents a novel approach to remote sensing image fusion. The core innovation is the FusionMamba block, an expansion of the Mamba block that can handle dual inputs to effectively merge spatial and spectral features with linear complexity. This block is incorporated into an interpretable network architecture that uses separate U-shaped branches with four-directional Mamba blocks to extract spatial and spectral features, merging them in an auxiliary combination branch with FusionMamba blocks, and an enhanced channel attention module for improved spectral representation. The method is the first application of SSM for hyper-spectral pansharpening and HISR tasks, and demonstrates superior performance on six datasets including WV3 and GF2 \cite{GID2020}. Though the approach is innovative, it has several limitations: The FusionMamba block is designed to support only two inputs, which restricts its use in fusion tasks that require more than two inputs. It requires both input feature maps to have the same number of channels, making it less flexible compared to concatenation operations. The performance of the FusionMamba block can also be sensitive to the selection of parameters, especially those associated with the state space model (SSM). While the FusionMamba block has linear computational complexity, it still has some computational overhead; a four-directional Mamba block has a FLOP count of 2HWD + 36HWCN, and a FusionMamba block has a FLOP count of 2HWD + 72HWCN, in contrast to a convolution layer, which has a FLOP count of 2HWD.

\begin{figure}[H]
                \centering
                \includegraphics[width=1\textwidth]{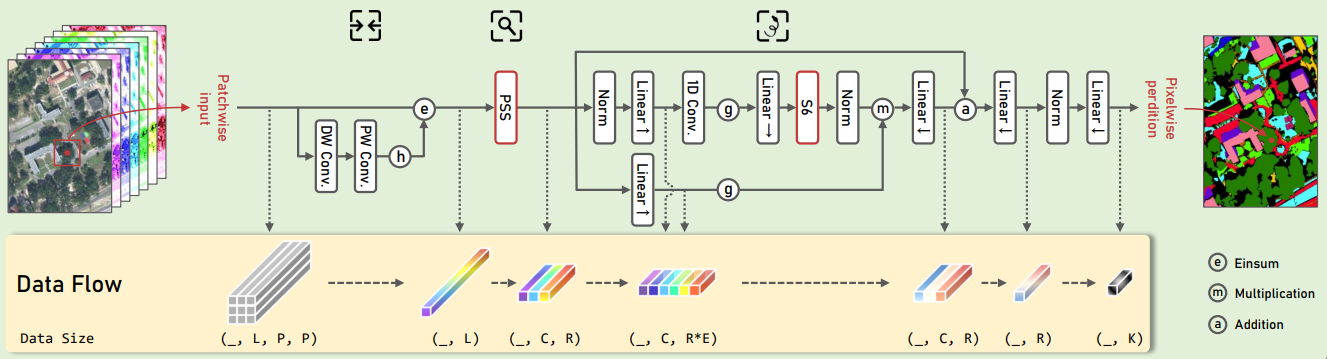}
                \caption{Spectral Mamba Architecture}
                \label{fig:spectral}
\end{figure}

 \textbf{SpectralMamba} \cite{yao2024spectralmambaefficientmambahyperspectral} presents a framework for hyperspectral (HS) image classification, integrating a state space model (SSM) to efficiently address spectral variability and redundancy. Its technical contributions include introducing Gated Spatial-Spectral Merging (GSSM) \cite{ahmad2024spatialspatialspectralmorphologicalmamba} to dynamically encode spatial-spectral relationships and mitigate spectral confusion, and Piece-wise Sequential Scanning (PSS), which segments long spectral sequences into manageable pieces, improving computational efficiency and preserving local spectral details. The model's architecture is built around an Efficient Selective State Space Model (S6). SpectralMamba addresses the curse of dimensionality, spectral variability, and the computational burden of existing models by using a simplified, lightweight architecture. SpectralMamba achieves superior performance across four benchmark datasets (Houston2013, Augsburg \cite{article}, Longkou, and Botswana), significantly outperforming conventional CNN, RNN, and Transformer models in overall accuracy (OA), average accuracy (AA), and the Kappa coefficient ($\kappa$), while maintaining resource efficiency. It has two major limitations: reliance on hyperparameter tuning (e.g., scanning piece size) and extreme spectral redundancy scenarios which suggest avenues for refinement.

\textbf{2DMamba} \cite{zhang20242dmambaefficientstatespace} is a novel 2D selective State Space Model (SSM) framework that addresses the limitations of applying 1D sequence models like Mamba to 2D image data. 2DMamba directly processes 2D images, maintaining spatial continuity, unlike 1D Mamba-based methods that lead to spatial discrepancies. This is achieved through a 2D selective scan operator, which performs parallel horizontal and vertical scans and is optimized for hardware using 2D tiling and caching to enhance computational efficiency. The 2DMamba architecture is validated on Giga-pixel Whole Slide Images (WSIs) \cite{Zhang_2022} and natural images. For WSIs, it uses a feature extractor for tissue patches and learnable tokens for non-tissue areas, and includes multiple 2DMamba blocks followed by an aggregator. The method shows strong performance across multiple WSI classification and survival analysis datasets. While 2DMamba maintains linear memory consumption, it experiences a decline in throughput for larger input sizes and is sensitive to parameter choices. The method does not benefit from positional embeddings, as its 2D formulation effectively captures spatial information. The back propagation process in 2DMamba requires four 2D selective scans, leading to high computational cost.

\begin{figure}[H]
                \centering
                \includegraphics[width=1\textwidth]{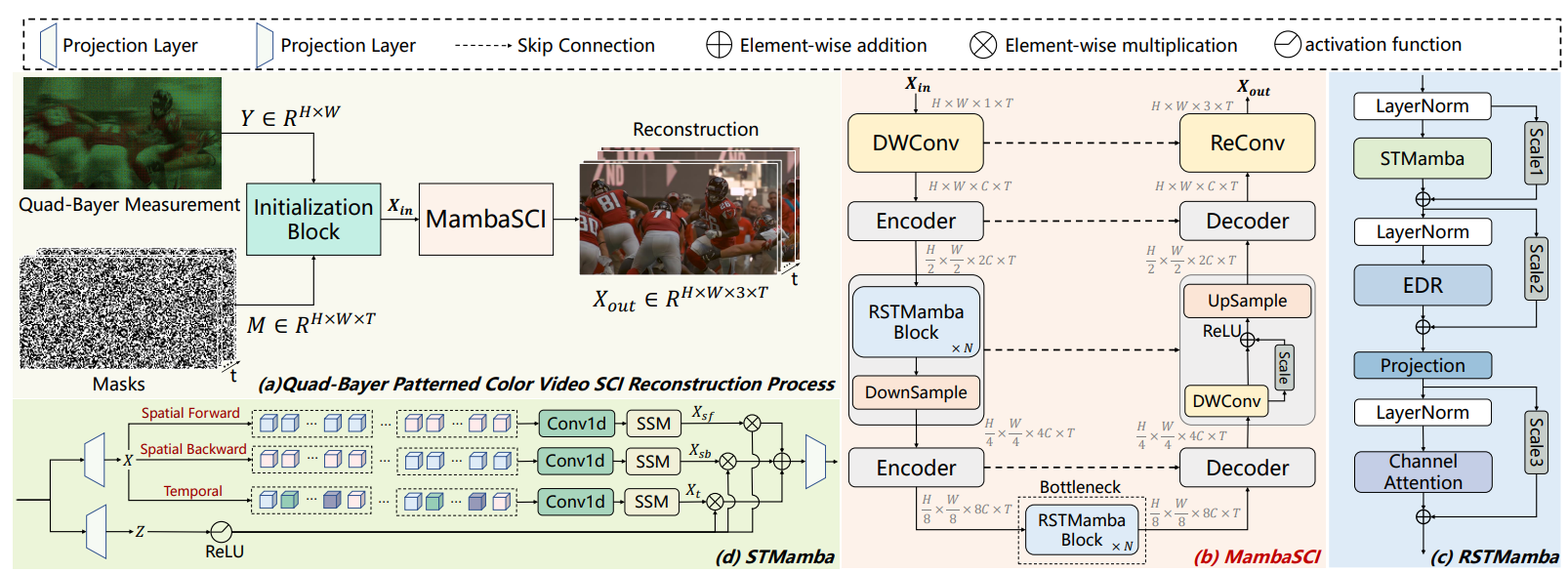}
                \caption{MambaSCI Architecture: Focuses on the reconstruction process for Quad-Bayer patterned color video. It includes projection layers, skip connections, element-wise operations, and activation functions. The architecture features STMamba and RSTMamba blocks for spatial and temporal processing, with layer normalization and control state space models (ControlSSM) for efficient data handling. The process involves encoding, decoding, and attention mechanisms to reconstruct high-quality video from patterned measurements, emphasizing the integration of spatial and temporal information for accurate reconstruction.}
                \label{fig:msci}
\end{figure}

\textbf{MambaSCI} \cite{pan2024mambasciefficientmambaunetquadbayer}  introduces the quad-Bayer color filter array (CFA) pattern into color video snapshot compressive imaging (SCI) \cite{9363502}, addressing the limitations of existing algorithms designed for traditional Bayer patterns. As the first method tailored for quad-Bayer patterns, it integrates the Mamba model with a non-symmetric UNet architecture \cite{ronneberger2015unetconvolutionalnetworksbiomedical}, customized for efficient and high-quality video reconstruction. The key technical contributions include the development of Residual-Mamba-Blocks, which combine Spatial-Temporal Mamba (STMamba) 
\cite{10681692} for capturing long-range spatial-temporal dependencies, Edge-Detail-Reconstruction (EDR) for recovering fine edge details lost during compression, and Channel Attention (CA) modules for enhancing inter-channel interactions and reducing artifacts. The network employs a hierarchical encoder-decoder design, incorporating residual connections and learnable scales to improve efficiency and reconstruction quality. At its core, the STMamba module captures spatial and temporal consistency using structured State-Space Models (SSMs) with linear complexity, while the EDR module integrates global and local information through depthwise convolutions, and the CA module dynamically weights features based on channel importance. MambaSCI is trained on the DAVIS2017 dataset \cite{ponttuset20182017davischallengevideo} and evaluated using middle- and large-scale simulation color video datasets. It outperforms state-of-the-art methods with fewer parameters and FLOPS, demonstrating better visual results. However, the model's limitations include a trade-off between computational complexity and performance due to the use of Conv3d in the CA module, and the lack of real quad-Bayer SCI data for real-world evaluations.

\textbf{ML-Mamba} \cite{huang2024mlmambaefficientmultimodallarge}, introduces a novel approach to multimodal learning by using the Mamba-2 state space model (SSM) as its backbone \cite{dao2024transformersssmsgeneralizedmodels}.It proposes a Mamba-2 Scan Connector (MSC), a multimodal connector that includes a Mamba-2 Visual Selective Scanning (MVSS) module and a SwiGLU module \cite{shazeer2020gluvariantsimprovetransformer}, to enhance representational capabilities when integrating visual and linguistic data. The model uses a pre-trained Mamba-2 language model, a visual encoder combining DINOv2 and SigLIP, and an MLP projector. The MSC module incorporates bidirectional and cross-scanning mechanisms to handle 2D visual data.ML-Mamba was trained on a dataset that includes a 558K subset of the LAION-CC-SBU dataset \cite{liu2024synthvlmhighefficiencyhighqualitysynthetic} for alignment and a 665K visual multi-round dialogue samples with text data for fine-tuning. ML-Mamba currently faces challenges in running on mobile devices due to its memory usage requirements. This is particularly relevant for deployment on resource-constrained devices like smartphones or tablets. It includes reliance on specific multimodal datasets which may introduce biases or incomplete coverage, this means that the model's performance might be suboptimal when faced with data that deviates significantly from its training set. \

\subsubsection{Hybrid Models}

\begin{figure}[H]
                \centering
                \includegraphics[width=1\textwidth]{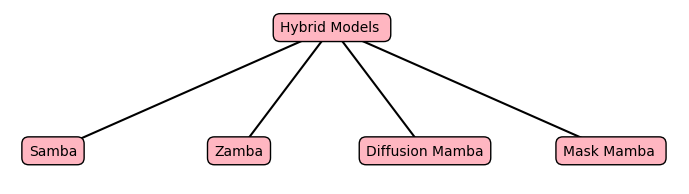}
                \caption{Models based on Hybrid Architecture}
                \label{fig:hymod}
\end{figure}

Hybrid architecture in small vision-language models (sVLMs) are designed to leverage the strengths of different architectural approaches, combining them to achieve efficient and effective processing of both visual and textual data. These models strive for a balance between computational efficiency and task performance, making them ideal for deployment in resource-constrained environments. By integrating the complementary features of vision and language processing components, hybrid models create lightweight yet powerful systems capable of cross-modal understanding and generation. A key advantage of hybrid architectures is their ability to use the most suitable approach for each modality. For instance, convolutional neural networks (CNNs) may be used for visual feature extraction, benefiting from their ability to capture spatial hierarchies in images. Simultaneously, transformers can be employed for sequence modeling, using their ability to manage complex relationships in textual data. Hybrid models also utilize cross-modal attention to seamlessly integrate information from different modalities. hybrid models in small VLMs offer a promising and flexible approach to vision-language understanding. They provide a versatile and scalable solution for various applications, allowing for practical use in real-world situations where resources are limited. As research progresses, these models are expected to further bridge the gap between performance and efficiency in vision-language tasks. The design of hybrid models allows for the integration of various strengths, providing a pathway to more effective multimodal AI systems.

\begin{figure}[H]
                \centering
                \includegraphics[width=1\textwidth]{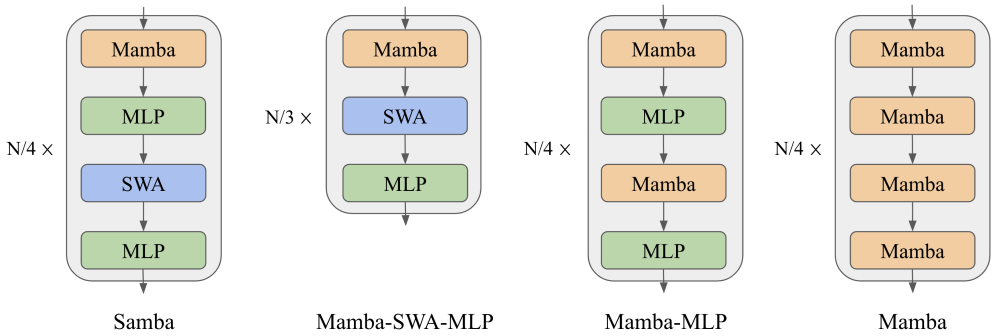}
                \caption{SAMBA: Illustrations of layer-wise integration of Mamba with various configurations of Multi-Layer Perceptrons (MLPs) and Sliding Window Attention (SWA).}
                \label{fig:samba}
\end{figure}

\textbf{SAMBA} \cite{ren2024sambasimplehybridstate}, a hybrid neural architecture that integrates selective State Space Models (SSMs) with attention mechanisms to address the challenges of efficiently modeling sequences with extremely long context lengths. SAMBA interleaves Mamba layers, Sliding Window Attention (SWA), and Multi-Layer Perceptrons (MLPs) to achieve linear computational complexity while effectively capturing both recurrent and non-recurrent dependencies. By leveraging the complementary strengths of SSMs for time-dependent semantics and SWA for precise memory retrieval, SAMBA outperforms state-of-the-art transformer-based models across short- and long-context benchmarks. It is trained on large-scale datasets such as SlimPajama \cite{shen2024slimpajamadcunderstandingdatacombinations}, comprising trillions of tokens, and shows strong extrapolation capabilities with minimal fine-tuning. While it comes with a lot of advantages, it still has some limitations which include the approaches required for the training from the scratch, the reliance on fine-tuning to enhance the recall of long contextual memory, and challenges in achieving robust zero-shot generalization for particular retrieval tasks. Moreover, these approaches frequently training from scratch or involve extra overhead, which makes it tough to apply them directly in pre-trained long-context language models.

\textbf{Zamba} \cite{glorioso2024zambacompact7bssm} introduces a state-of-the-art-transformer-SSM hybrid architecture at 7
Billion parameters, maximizing flop efficiency as well as the attention-based in context learning. It also introduced a neuroscience-inspired shared attention optimization that preserves performance while reducing memory usage. It reduces the quadratic cost of the core attention computation in transformers which is the bottleneck of this architecture and also it enhances the performance of the early SSM models on language tasks. Zamba features a global shared attention architecture, reusing mutual parameters every six MAMBA blocks to enhance FLOP usage and performance. MMLU \cite{hendrycks2021measuringmassivemultitasklanguage}, Human Eval \cite{chen2021codex}, GSM8k \cite{cobbe2021gsm8k}, ARC \cite{allenai:arc}, HellaSwag \cite{zellers2019hellaswag}, Winogrand \cite{ai2:winogrande}, and PIQA zero-shot QA \cite{Bisk2020} are some of the datasets. There is a phase 1 dataset which includes the pile and refined web, whereas annealing datasets have a large collection of existing high-quality sources. It matches with the other state-of-art-of models in different linguistic evaluations, however, it faces some of the limitations which include lagging performance in the tests of reasoning and in context learning, code generation due to a lack of specialized data, reliance on fewer tokens compared to competing models, and the need for more extensive fine-tuning to close gaps with state-of-the-art models. One of the biggest drawbacks of this architecture is that the quadratic cost of attentional mechanisms in the transformer related to sequence length increases leading to the quadratic growth of the amount and computation of the memory required. which shows that the larger sequences significantly increase the resources requirements thus creating a challenge to handle large contexts efficiently.If overcoming this limitation is possible then it  is possible that it would enhance the context lengths and much more efficient generation in the term of both the FLOPs and the memory required to store the KV cache.

\begin{figure}[H]
                \centering
                \includegraphics[width=1\textwidth]{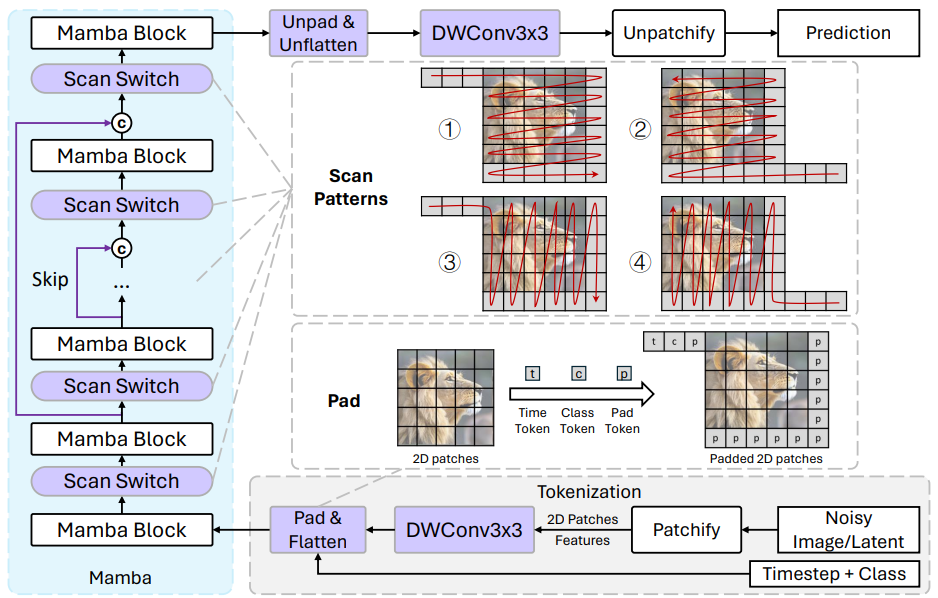}
                \caption{Diffusion Mamba (DiM) Architecture: The inputs of the framework is a noisy image/latent, with a timestep and a
class condition. The noisy inputs are transformed into patch-wise features, processed by a depth-wise
convolution, and appended with time, class, and padding tokens. The features are flattened and
scanned by Mamba blocks with four directions. The features are then transformed into 2D patches,
processed by another convolution, and finally used for noise prediction.}
                \label{fig:dim}
\end{figure}

\textbf{Diffusion Mamba (DiM)} \cite{teng2024dimdiffusionmambaefficient} is a state-of-the-art diffusion model that integrates the Mamba sequence model to address computational inefficiencies in high-resolution image synthesis. DiM innovatively adapts Mamba, a model originally designed for 1D sequential
data, to 2D image generation by introducing multi-directional scanning patterns, learn able padding tokens to maintain spatial coherence, and lightweight local feature enhance ment via depth-wise convolutions. DiM employs training-free upsampling techniques, enabling it to generate ultra-high-resolution images (up to 1536×1536) without additional fine-tuning. Evaluations on ImageNet \cite{russakovsky2015imagenetlargescalevisual} and CIFAR-10 \cite{sun2024cifar10warehousebroadrealistictestbeds} datasets demonstrate its competitive performance. However, DiM has limitations where the first one is the cost as training the Mamba-based diffusion models on the high-resolution images is extremely costly.Apart from it faces some more limitations including difficulties in maintaining spatial coherence in high-resolution human image generation, instability in fine details such as facial features and limb positions, and challenges with repetitive patterns during training-free upsampling at extreme resolutions. The model’s training process remains resource-intensive for very high-resolution tasks, and its effectiveness is dependent on pretraining quality and dataset diversity, potentially limiting its adaptability to highly specific scenarios.

\begin{figure}[H]
                \centering
                \includegraphics[width=1\textwidth]{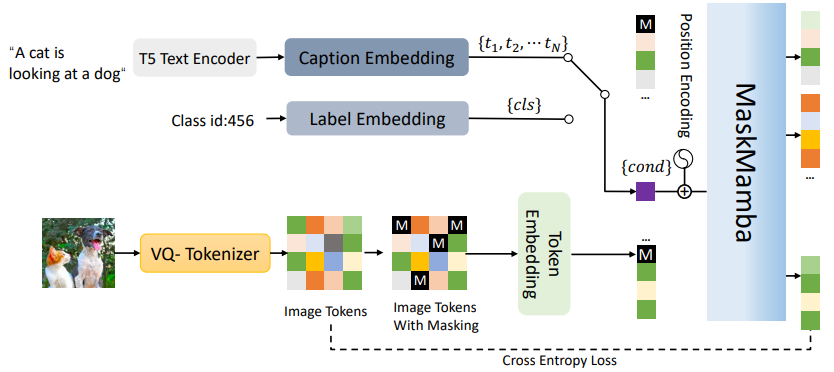}
                \caption{MaskMamba Architecture: Integrates text and image processing for tasks like image captioning. It uses a T5 text encoder to generate caption embeddings and a VQ-Tokenizer to process image tokens, incorporating masking techniques. The architecture includes position encoding and cross-entropy loss to align text and image representations effectively. This design enables the model to handle multimodal data, ensuring accurate and context-aware image caption generation.}
                \label{fig:mask}
\end{figure}

\textbf{MaskMamba} \cite{chen2024maskmambahybridmambatransformermodel} is a hybrid generative model that integrates Mamba and Transformer
architectures to enable efficient, non-autoregressive masked image generation. The model introduces several architectural innovations, including a redesigned Bi-Mamba layer that replaces causal convolutions with standard convolutions to enhance bidirectional token in interactions and global context modeling. MaskMamba incorporates hybrid configurations, such as grouped parallel and serial mixing schemes, to optimize feature representation and scalability. It addresses key challenges in image generation, including the quadratic complexity of Transformers and the inefficiencies of autoregressive approaches, by leveraging
Masked Image Modeling for faster decoding with minimal computational overhead. Evaluations on ImageNet \cite{russakovsky2015imagenetlargescalevisual} and CC2M \cite{lin2015microsoftcococommonobjects} datasets demonstrate MaskMamba’s superior performance, achieving high generation quality and a 54.44\% improvement in inference speed at 2048×2048 resolution compared to Transformer-based baselines. MaskMamba faces limitations, including sensitivity to the placement of condition embeddings, dependency on high-quality text prompts for optimal text-to-image synthesis, and challenges in scaling with limited or noisy training data. These constraints may impact its generalization and adaptability to diverse tasks and dataset.

\section{Evaluation}

The presented models exhibit a diverse range of architectures, primarily encompassing Transformer-based, Hybrid (CNN + Transformer), and Mamba models. These architectures are evaluated across a variety of benchmarks which can be seen in the Table \ref{t1}, including tasks such as Visual Question Answering (VQA), Generalized Question Answering (GQA), TextVQA, and multimodal applications like MM-Vet and MMBench. The performance metrics span accuracy, mIoU, PSNR, AUC, and FLOPS, with models like TinyViT, MiniGPT, and FastViT demonstrating efficiency through reduced parameter sizes, while maintaining competitive performance. Larger models such as Cobra-8B and Falcon-Mamba-7B generally achieve high accuracy, albeit at the cost of substantial computational demands, which may limit their applicability in resource-constrained environments. A noticeable trend across the models is the emphasis on reducing parameter sizes and improving inference speed, as seen with the LLaVA and FastVLM series, which achieve high performance on various tasks while maintaining reduced resource footprints. Specialized models like Mini-Gemini and 2DMambaMIL excel in their specific domains, such as multimodal learning and medical image analysis, offering impressive results despite smaller model sizes. Models from the Mamba family, including SpectralMamba and MambaSCI, showcase significant improvements in task-specific performance, particularly in hyperspectral image classification and multivariate time series forecasting, with markedly lower computational resource consumption. Overall, the models highlight a growing focus on optimizing efficiency without compromising accuracy, leveraging hybrid architectures and attention mechanisms to achieve superior results in both standard and domain-specific benchmarks.

\begin{table}[H]
\centering
\caption{Benchmarking the Models}
\label{t1}
\resizebox{\textwidth}{!}{
\begin{tabular}{|l|l|l|l|l|l|}
\hline
\textbf{Model Name} & \textbf{Architecture} & \textbf{Parameter Size} & \textbf{Key Benchmarks} & \textbf{Key Metrics} & \textbf{Results} \\ \hline
TinyViT-21M & Vision Transformer & 21M & ImageNet-21k & Top-1 Accuracy & 84.8 (IN-1k) \\ \hline
TinyViT-21M (Hybrid) & Hybrid & 21M & IN-1k & Top-1 Accuracy & 86.2 (384), 86.5 (512) \\ \hline
FastViT-MA36 & Transformer & N/A & ADE20k & mIoU & 5.2\% higher than PoolFormer-M36 \\ \hline
MiniGPT-4 & Transformer & N/A & Meme, Recipe, Ad, Poem & Success Rate & ~80\% (Recipes, Ads, Poems); 8/25 (Memes) \\ \hline
TinyGPT-V & Transformer & 2.8B & VSR, GQA & VSR Acc., GQA Score & 54.7\% (VSR), 38.9\% (GQA) \\ \hline
TinyGPT-V & Transformer & 2.8B & VQA & Sec./word, Inf. Occ. & 0.067 s/word, 5.6GB \\ \hline
Mini-Gemini & Transformer & N/A & VQAT, MMB, MME, MM-Vet & Various & Outperforms existing models \\ \hline
Mini-Gemini-HD & Transformer & N/A & TextVQA & Accuracy & 74.1\% (Hermes-2-Yi-34B) \\ \hline
LLaVA-1.5 (Baseline) & N/A & N/A & TextVQA, MME, MM-Vet & Accuracy & 54.1, 1467.1, 30.7 \\ \hline
LLaVA-1.5 (+ Info mining) & N/A & N/A & TextVQA, MME, MM-Vet & Accuracy & 58.1, 1485.2, 31.3 \\ \hline
LLaVA-1.5 (+ Higher res.) & N/A & N/A & TextVQA, MME, MM-Vet & Accuracy & 59.8, 1478.3, 31.9 \\ \hline
LLaVA-1.5 (Baseline) & N/A & N/A & TextVQA, MME, MM-Vet & Accuracy & 58.2, 1510.7, 31.1 \\ \hline
LLaVA-1.5 (+ Info mining) & N/A & N/A & TextVQA, MME, MM-Vet & Accuracy & 58.4, 1451.7, 33.8 \\ \hline
LLaVA-1.5 (+ Larger VE-HR) & N/A & N/A & TextVQA, MME, MM-Vet & Accuracy & 61.5, 1517.0, 34.6 \\ \hline
FastViTHD & Hybrid (CNN + Transformer) & N/A & GQA, TextVQA, POPE, DocVQA, SeedBench & Avg-5 VLMs & 54.4 (256x256), 62.6 (768x768) \\ \hline
FastVLM & Hybrid (CNN + Transformer) & N/A & GQA, TextVQA, POPE, SeedBench & TTFT, Avg-5 VLMs & 3.2x faster TTFT, 7.9x faster than Cambrian-1 \\ \hline
FastVLM (R3) & Hybrid (CNN + Transformer) & N/A & GQA, SQA, TextVQA, POPE, LLaVA, MM-Vet, VQAv2, DocVQA, SeedBench & TTFT, VLM Eval. & 166ms, 61.6\%, 61.4\%, 57.4\%, 87.4\%, 56.0\%, 31.8\%, 77.0\%, 61.0\%, 30.9\%, 65.6\% \\ \hline
FastVLM (R4) & Hybrid (CNN + Transformer) & N/A & GQA, SQA, TextVQA, POPE, LLaVA, MM-Vet, VQAv2, DocVQA, SeedBench & TTFT, VLM Eval. & 166ms, 62.9\%, 80.8\%, 61.6\%, 87.0\%, 63.6\%, 31.0\%, 78.7\%, 66.2\%, 31.7\%, 68.9\% \\ \hline
FastVLM (R38) & Hybrid (CNN + Transformer) & N/A & GQA, SQA, TextVQA, POPE, LLaVA, MM-Vet, VQAv2, DocVQA, SeedBench & TTFT, VLM Eval. & 125ms, Outperforms Cambrian-1 (7.9x faster) \\ \hline
FastVLM (R39) & Hybrid (CNN + Transformer) & N/A & GQA, SQA, TextVQA, POPE, LLaVA, MM-Vet, VQAv2, DocVQA, SeedBench & TTFT, VLM Eval. & 125ms, Beats Cambrian-1 (2.3x fewer tokens) \\ \hline
FastVLM (R3) & Hybrid (CNN + Transformer) & N/A & ChartQA, OCRBench, TextVQA, DocVQA, InfoVQA & TTFT, Accuracy & 446ms, 69.3\%, 45.9\%, 69.5\%, 66.9\%, 34.3\% \\ \hline
FastVLM (R8) & Hybrid (CNN + Transformer) & N/A & ChartQA, OCRBench, TextVQA, DocVQA, InfoVQA & TTFT, Accuracy & 446ms, 74.2\%, 59.0\%, 72.8\%, 72.0\%, 44.3\% \\ \hline
LLaVA-Mini & Transformer & N/A & VQAv2, GQA, MMBench & Accuracy & Comparable to LLaVA-v1.5 (1 vision token) \\ \hline
LLaVA-Mini (1fps) & Transformer & N/A & EgoSchema & Accuracy & 51.2 \\ \hline
LLaVA-Mini (w/o compression) & Transformer & N/A & VQAv2, GQA, MMB & Accuracy & 80.0, 62.9, 66.2 \\ \hline
2DMambaMIL & Mamba & N/A & PANDA, TCGA-BRCA & Acc., AUC & 0.5075, 0.8184 (PANDA); 0.9458, 0.9782 (TCGA-BRCA) \\ \hline
Cobra-8B & Mamba & 8B & VQA, VizWiz, VSR & Avg. Accuracy & Best across all benchmarks \\ \hline
Cobra-3.5B & Mamba & 3.5B & VQA, VizWiz, VSR & Avg. Accuracy & 77.8, 62.3, 49.7, 58.2, 58.4, 88.4 \\ \hline
Falcon-Mamba-7B & Mamba & 7B & MMLU, GSM8K, TruthfulQA & Avg. Score & Outperforms similar-scale models \\ \hline
FusionMamba & Mamba & 0.73M & Pansharpening, Hyper-spectral & PSNR, Q2n, etc. & Best on WV3, GF2 datasets \\ \hline
MambaSCI & Mamba & 6.11M & Color video recon. & PSNR, SSIM & Outperforms EfficientSCI \\ \hline
ML-Mamba & Mamba-2 & 3B & VQA-v2, GQA, VQAT, POPE, VizWiz, VSR & Accuracy & Comparable to LLaVA v1.5 (40\% fewer params) \\ \hline
SiMBA & Hybrid (Mamba + EinFFT) & N/A & Time Series, Image Class. & Top-1 Accuracy & Bridges gap between SSMs and Transformers \\ \hline
SpectralMamba & Mamba & N/A & Hyperspectral Image Class. & OA & Outperforms competitors \\ \hline
SpectralMamba (patchwise) & Mamba & N/A & Forest, Residential, Industrial & OA & 98.31, 73.79, 48.71 \\ \hline
VL-Mamba & Mamba & 2.8B & VQA-v2, GQA, SQAI, VQAT, POPE, MME, MMB, MM-Vet & Accuracy & 74.5, 54.4, 63.4, 44.6, 84.9, 1381.8, 55.8, 30.6 \\ \hline
Zamba-7B (Phase 1) & Mamba & 7B & MMLU & MMLU & 50.82 \\ \hline
Zamba-7B (Annealed) & Mamba & 7B & MMLU & MMLU & 57.72 \\ \hline
Samba-3.8B & Hybrid (Mamba + Attention) & 3.8B & MMLU, GSM8K, HumanEval & Avg. Score & Highest among pre-trained models \\ \hline
Samba-3.8B & Hybrid (Mamba + Attention) & 3.8B & GSM8K & Accuracy & 18.1\% higher than TFM++ \\ \hline
MaskMamba-XL & Hybrid (Mamba + Transformer) & 741M & ImageNet & FID-50k & 5.79 \\ \hline
\end{tabular}
}
\end{table}

\section{Advancing Research in Small Vision-Language Models}  

The domain of small vision-language models (sVLMs) offers a diverse range of avenues for research and development, with significant potential to drive innovations across computational efficiency, architecture design, and multimodal capabilities. To further propel advancements in this field, the following areas present critical opportunities for exploration:  

\begin{enumerate}  
    \item \textbf{Data Efficiency and Synthetic Data Generation:}  
    Reducing reliance on large-scale pretraining datasets is a key challenge for sVLMs. Techniques such as synthetic data generation, domain-specific augmentation, and unsupervised or self-supervised learning could allow models to excel in data-scarce environments. Leveraging generative adversarial networks (GANs) \cite{goodfellow2014generativeadversarialnetworks} and diffusion models for synthetic dataset creation tailored to specific applications, such as medical imaging or autonomous systems, can enhance performance without extensive labeled data.  

    \item \textbf{Cross-Domain and Task Generalization:}  
    Ensuring sVLMs generalize effectively across diverse domains and complex tasks is essential. Research into modular architectures capable of adapting to domain-specific nuances through meta-learning or task-conditioned pretraining could bridge the gap between general-purpose and specialized performance. Developing methodologies to align representations across domains, particularly for emerging applications like remote sensing, robotics, and document understanding, remains a promising direction.  

    \item \textbf{Advanced Architectures for Edge Devices:}  
    The computational constraints of edge devices and IoT platforms necessitate innovative architecture designs. Research into ultra-low-power algorithms and lightweight attention mechanisms, such as sparse attention and low-rank matrix approximations, can minimize memory and latency costs. Exploring state-space models (SSMs) \cite{alonso2024statespacemodelsfoundation} and hybrid transformer-based designs with hardware-aware optimizations for devices like FPGAs \cite{article} or TPUs \cite{jouppi2023tpuv4opticallyreconfigurable} could redefine the efficiency landscape for sVLMs.  

    \item \textbf{Multimodal Integration Beyond Text and Vision:}  
    Expanding sVLMs to support modalities such as audio, haptics, and sensor data provides an opportunity to create truly comprehensive AI systems. For instance, integrating temporal and spatial sensor information alongside visual and textual data could enhance applications in autonomous systems, human-computer interaction, and augmented reality. Techniques for modality alignment and fusion, including dynamic cross-modal attention and hierarchical embeddings, warrant further investigation.  

    \item \textbf{Bias Mitigation and Fairness:}  
    Addressing biases in training data and improving fairness in sVLM outputs is critical for ensuring equitable applications. Robust methodologies for identifying and mitigating biases—such as adversarial debiasing, fairness-aware training, and interpretability-focused evaluations—can improve model trustworthiness, especially in sensitive domains like hiring, healthcare, and law enforcement.  

    \item \textbf{Explainability and Interpretability:}  
    Transparent decision-making processes are increasingly crucial for sVLM applications in critical fields. Research into interpretable architectures, attention visualization, and layer-wise relevance propagation could help demystify model behaviors. Additionally, integrating explainability as a first-class objective during pretraining and fine-tuning phases could yield architectures inherently aligned with interpretability.  

    \item \textbf{Scalable Training Strategies:}  
    Efficient training methodologies, such as curriculum learning, progressive training, and parameter-efficient fine-tuning (e.g., LoRA\cite{hu2021loralowrankadaptationlarge} or adapters), offer significant potential to enhance scalability. Distributed training strategies that balance memory efficiency and computational speed can enable researchers to train larger models even in constrained environments.  

    \item \textbf{Emerging Evaluation Paradigms:}  
    Current benchmarks may inadequately reflect real-world performance. Developing nuanced evaluation metrics that capture multimodal alignment, robustness to noisy data, and downstream task relevance is a vital research direction. In particular, evaluating models under adversarial scenarios or real-time conditions can better guide practical implementations.  

    \item \textbf{Hybrid Architectures and Model Compression:}  
    The integration of state-space models (SSMs) \cite{alonso2024statespacemodelsfoundation} with transformers, convolutional layers, and recurrent mechanisms can offer a balanced approach to scalability, efficiency, and performance. Research into advanced model compression techniques, such as knowledge distillation, pruning, and quantization, can enable compact models to achieve near-parity with larger architectures. These approaches are particularly relevant for scaling sVLMs in real-time applications and memory-constrained environments.  

    \item \textbf{Cross-Language Multimodal Learning:}  
    Leveraging multilingual data for both text and image components can make sVLMs more inclusive and versatile. Innovations in joint multilingual embeddings, transfer learning across linguistic domains, and cross-language pretraining objectives could allow for more robust cross-cultural AI systems. These systems could support applications ranging from global healthcare to multilingual education platforms.  

    \item \textbf{Unexplored Applications of sVLMs:}  
    While sVLMs have seen applications in visual question answering and image captioning, new domains remain underexplored. For example, adapting sVLMs for scientific discovery, such as protein-structure prediction or climate modeling, presents a promising avenue. Similarly, integrating sVLMs into interactive AI agents for educational or assistive purposes could address unique challenges requiring multimodal interaction.  

    \item \textbf{Incorporating Continual Learning:}  
    Enabling sVLMs to learn incrementally without catastrophic forgetting is a promising research area. Techniques such as dynamic network expansion, replay mechanisms, and task-specific regularization could empower sVLMs to adapt seamlessly to new tasks or domains without significant retraining.  
\end{enumerate}  

By pursuing these directions, researchers can address existing limitations and unlock the transformative potential of sVLMs across a wide array of applications. These opportunities not only promise advancements in efficiency and scalability but also underscore the broader implications of accessible and versatile AI systems in diverse societal contexts.

\section{Conclusion}
The development of small vision-language models (sVLMs) marks a pivotal step towards creating accessible, efficient, and versatile multimodal AI systems. This survey has highlighted the evolution from early models like CLIP to advanced architectures such as VL-Mamba \cite{qiao2024vlmambaexploringstatespace}, reflecting a trend towards achieving computational efficiency without compromising performance. We explored the significance of transformer-based models, knowledge distillation, and lightweight attention mechanisms in shaping sVLMs. Key findings include the performance and efficiency trade-offs observed in models like TinyGPT-V \cite{yuan2024tinygptvefficientmultimodallarge}, MiniGPT-4 \cite{zhu2023minigpt4enhancingvisionlanguageunderstanding}, and FastVLM \cite{vasu2024fastvlmefficientvisionencoding}. The emergence of Mamba-based architectures presents an alternative to transformers \cite{vaswani2023attentionneed}, emphasizing scalability and resource optimization. Hybrid models further demonstrate the potential for balancing performance and efficiency through cross-modal integration. Despite significant advancements, challenges remain in addressing dataset biases, improving generalization, and enhancing performance in complex tasks. Future research should focus on robust architectural innovations, refined pre-training strategies, and evaluation metrics that better reflect real-world applications. The rapid evolution of sVLMs underscores the importance of this survey in offering a comprehensive overview of the current state and future directions in the field, inspiring continued progress toward efficient multimodal AI.

\bibliographystyle{plain}
\bibliography{main}

\end{document}